# DESIGN AND IMPLEMENTATION OF REAL-TIME ALGORITHMS FOR EYE TRACKING AND PERCLOS MEASUREMENT FOR ON BOARD ESTIMATION OF ALERTNESS OF DRIVERS

*Anjith George*

# DESIGN AND IMPLEMENTATION OF REAL-TIME ALGORITHMS FOR EYE TRACKING AND PERCLOS MEASUREMENT FOR ON BOARD ESTIMATION OF ALERTNESS OF DRIVERS

*Report submitted to*
*Indian Institute of Technology, Kharagpur*
*for the award of the degree*

*of*

**Master of Technology**
**in Electrical Engineering with Specialization**
**in "Instrumentation"**

*by*

**Anjith George**

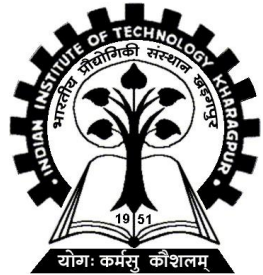

**DEPARTMENT OF ELECTRICAL ENGINEERING**
**INDIAN INSTITUTE OF TECHNOLOGY, KHARAGPUR**
**April 2012**

# DECLARATION

I certify that

a. the work contained in this report is original and has been done by me under the guidance of my supervisor(s).
b. the work has not been submitted to any other Institute for any degree or diploma.
c. I have followed the guidelines provided by the Institute in preparing the report.
d. I have conformed to the norms and guidelines given in the Ethical Code of Conduct of the Institute.
e. whenever I have used materials (data, theoretical analysis, figures, and text) from other sources, I have given due credit to them by citing them in the text of the report and giving their details in the references. Further, I have taken permission from the copyright owners of the sources, whenever necessary.

Signature of the Student

# CERTIFICATE

This is to certify that the **Dissertation Report** entitled, "**Design and Implementation of Real Time Algorithms for Eye Tracking and PERCLOS measurement for On board Estimation of Alertness of Drivers**" submitted by **Mr. "Anjith George"** to Indian Institute of Technology, Kharagpur, India, is a record of bonafide Project work carried out by him under my supervision and guidance and is worthy of consideration for the award of the degree of Master of Technology in Electrical Engineering with Specialization in "**Instrumentation**" of the Institute.

…………………………….

Date:

(Supervisor)
Prof. Aurobinda Routray
Department of Electrical Engineering
Indian Institute of Technology, Kharagpur

iii

# ACKNOWLEDGEMENTS

This thesis is the result of research performed under the guidance of Prof Aurobinda Routray at the Department of Electrical Engineering of the Indian Institute of Technology, Kharagpur.

I am deeply grateful to my supervisor for having given me the opportunity of working as part of his research group and huge amount of time and effort he spent guiding me through several difficulties on the way. Without the help, encouragement and patient support I received from my guide, this thesis would never have materialized

I also acknowledge the research scholars, Anirban Dasgupta, Tapan Pradhan and S.L. Happy for their constant encouraging suggestions, sharing the technical skills and the synergy that they brought in my work. Further I would like to thank my friends and my parents for their wholehearted supports during the work.

Anjith George

Department of Electrical Engineering

Indian Institute of Technology, Kharagpur

April 2012

# ABSTRACT

The alertness level of drivers can be estimated with the use of computer vision based methods. The level of fatigue can be found from the value of PERCLOS. It is the ratio of closed eye frames to the total frames processed. The main objective of the thesis is the design and implementation of real-time algorithms for measurement of PERCLOS. In this work we have developed a real-time system which is able to process the video onboard and to alarm the driver in case the driver is in alert. For accurate estimation of PERCLOS the frame rate should be greater than 4 and accuracy should be greater than 90%. For eye detection we have used mainly two approaches Haar classifier based method and Principal Component Analysis (PCA) based method for day time. During night time active Near Infra Red (NIR) illumination is used. Local Binary Pattern (LBP) histogram based method is used for the detection of eyes at night time. The accuracy rate of the algorithms was found to be more than 90% at frame rates more than 5 fps which was suitable for the application.

**Keywords**: PERCLOS, Face detection, Real-time system, Local binary pattern.

# CONTENTS

# LIST OF ABBREVIATIONS

**PERCLOS**- Percentage closure of eye

**NIR**- Near infra red

**ROI**- Region of interest

**PCA**- Principal Component Analysis

**SVM** – Support Vector Machines

**LBP**-Local Binary Pattern

**SF**- Scale factor

**SBC**- Single board computer

# LIST OF FIGURES

# Chapter 1

# INTRODUCTION

The lack of alertness is the prime reason for a number of motor accidents reported [1-2]. The measurement of the fatigue level is crucial in avoiding the road causalities. Percentage closure of eyes (PERCLOS) is reported as a good indicator for fatigue detection [3]. It is the ratio of closed eye frames to the total frames processed. The level of fatigue can be found from the value of PERCLOS.

PERCLOS can be measured using non intrusive computer vision methods. The camera picks up the video and the videos are processed in real-time to give the PERCLOS values. The values are obtained in a one minute window and alarm is turned on if it is more than preset threshold.

The main challenges of real time and on board estimation of PERCLOS are that the algorithm should exhibit good accuracy with illumination variation, changing backgrounds and vibration of camera. For accurate estimation of PERCLOS the frame rate should be greater than 4 fps. For capturing video under night driving conditions an active Near Infra Red (NIR) lighting system is used.

The main focus is the design and implementation of image processing algorithms in a real-time embedded platform. We chose to implement the algorithms in Intel Atom Single board computer. The operating RTOS is chosen to be the Windows Embedded XP. Resolution of web cam is 640x480. For accurate estimate of PERCLOS the processing speed needs to be at least 4 frames per second. Since the system is to be installed in a vehicle, the algorithm has to be tolerant to illumination variations, different face poses and vibration of camera.

## 1.1 Literature Survey

The main approaches for eye detection and tracking is classified into two: they are active IR (infrared) based methods and passive appearance methods. Active IR illumination methods utilize a special bright pupil effect. It is an effective and simple approach for easy detection and tracking of eyes. The method works on a differential infra red scheme [4].In this method Infrared sources of two frequencies are used. First image is captured with infrared lighting at 850 nm, it produces a distinct glow in the pupils (the red-eye effect).The second image uses a 950nm infrared source for illumination that produces an image with dark pupils. These two images are synchronous with the camera and differ only by the brightness of pupil region. Now the difference of the two images are found in which the pupil region will be highlighted. After post processing the pupil blobs are identified and used for the tracking of eyes [5].

The main problems with this method are that the success rate changes with several factors. Brightness and size of pupils face orientation, external light interference (mainly the light from other vehicles and street lights) distance of driver from camera. The intensity of external light should be limited. The reflection and glints with glasses is another problem. Recently many developments are done in tuning the irradiation of IR illuminators. IR illuminators have to be tuned in order to operate in different natural light conditions, multiple reflections of glasses, and variable gaze directions. Some authors tried to implement systems which combines the active IR methods with appearance based methods. By combining imaging by using IR light and appearance-based object recognition techniques, those methods can robustly track eyes even when the pupils are not very bright due to significant external illumination interferences. The appearance model is incorporated in both eye detection and tracking via the use of a support vector machine and mean shift tracking [6].

Appearance based methods generally consists of two steps: locating face to extract eye regions and then eye detection from eye windows.  The face detection problem has been faced up with different approaches: neural network, principal components, independent components, and skin colour based method. Each of the methods is having some constraints: frontal view, expressionless images, limited variations of light conditions, uniform background, face colour variations and so on. Eriksson and Papanikolopoulos present a system to locate and track eyes of the driver. They use symmetry-based approach to locate the face in grayscale image, and then eyes are found and tracked.

Template matching is used to determine if the eyes are open or closed [7]. Singh and Papanikolopoulos proposed a non-intrusive vision-based system for the detection of driver fatigue [8]. The system uses a colour video camera that points directly towards the driver's face and monitors the driver's eyes in order to detect micro-sleeps.

Paul Viola and Michael Jones proposed a method for face detection using Haar like feature. The Haar like features can be computed fast using integral images. The algorithm is trained using AdaBoost method [9]. A strong classifier is formed by the combination of a large number of weak classifiers. The detector works in a cascaded manner so that the faces like regions passed by the first stages of classifier are more intensively processed. The detection rate of this method was more than 95% and the computational complexity is less. Turk and Pentland proposed Eigen face based method for face detection [10]. It included training with a large number of face images and the principal components are found. In the detection phase extensive block matching is done, with the projection of each block into the Eigen space. The reconstruction error is found for each block and the minimum value is found. If the minimum value is less than a threshold, it is considered as a positive detection. The major disadvantages were the variation with rotation, translation, size variation and illumination variations.

In some approaches authors used colour, edge, and binary information to detect eye pair candidate regions from input image, then extract face candidate region with the detected eye pair. They verified both eye pair candidate region and face candidate region using Support Vector Machines (SVM) [11]. Some methods use the appearance and dimensions of eye. Many eye tracking applications only need the detection and tracking of either the iris or the pupil. Depending on the viewing angle, both the iris and pupil appear elliptical and consequently can be modelled by five shape parameters. The Hough transform can be used effectively to extract the iris or the pupil, but requires explicit feature detection. Often a circularity shape constraint is employed for efficiency reasons, and consequently, the model only works on near-frontal faces.

Work on driver fatigue detection, has yielded many driver monitoring systems. All of these systems focus on providing information to drivers that will facilitate their driving and increase traffic safety. Some of them are given below

DAISY (Driver AssIsting System) has been developed asa monitoring and warning aid for the driver in longitudinal and lateral control on German motorways [12]. The warnings are generated based on the knowledge of behavioural state and condition of the driver.

Robotics Institute in Carnegie Mellon University developed a drowsy driver monitor Copilot [13]. The Copilot is a video-based system for measuring slow eyelid closure as represented by PERCLOS.

DAS (Driver Assistance System) has been developed by the group at the Australian National University [14]. It uses a dashboard-mounted faceLAB head-and-eye-tracking system to monitor the driver, The Distillation algorithm is used to monitor driver performance. Feedback on deviation in lane tracking is provided to the driver using force feedback to the steering wheel which is proportional to the amount of lateral offset estimation by the lane tracker.

### 1.2 Scope Of thesis

In the current work we consider the detection and classification of eyes in three stages

　　i.　Face detection and eye localization
　ii.　Eye detection
　iii.　Eye state classification

The false positive rate of algorithms is reduced by this approach and the search region for computationally intensive stages also reduces. For face detection we have selected Haar classifier based approach owing to its high accuracy rate [15]. The algorithm is modified for improved frame rate and detection of tilted faces. During daytime Haar based and PCA based algorithms are used for eye detection. In night time LBP (Local Binary pattern) based features are used for eye detection. Finally the state of eye is classified into open or closed with Support Vector Machine (SVM) [11] approach. The PERCLOS values are found over one minute windows.

## 1.3 Thesis Outline

Chapter 2 includes the requirement analysis of the system. The software and hardware requirements and the specifications of the system selected are discussed in detail.

Chapter 3 discusses the face detection algorithm and its modification for the improvement of real-time performance. An analysis of the variation of accuracy with different scale factors is also presented. It also discusses the detection of tilted faces.

Chapter 4 includes the eye detection methods for daytime. Haar classifier based method and Principal Component analysis based methods are discussed here.

Chapter 5 presents Local binary pattern features and their use in the detection of eyes in NIR lighting,

Chapter 6 discusses the eye state classification with SVM

Chapter 7 includes the details of the onboard testing of the system.

Chapter 8 includes Discussions and conclusions

# Chapter 2

# REQUIREMENT ANALYSIS

Real-time computing (RTC) is the study of hardware and software systems that are subject to a "real-time constraint"—i.e., operational deadlines from event to system response. A real time system may be one where its application can be considered (within context) to be mission critical. A system is said to be *real-time* if the total correctness of an operation depends not only upon its logical correctness, but also upon the time in which it is performed. Real time systems can be classified into 2 types

  i. Hard Real Time Systems
 ii. Soft Real Time Systems

A hard real time system is one where the completion of an operation after its deadline is considered to be useless – maybe leading to a critical failure of the entire system. Hard real-time systems are used when it is imperative that an event is reacted to within a strict deadline [15]. For example, a pacemaker is an example of a hard real time system. Where as a soft real-time system will tolerate such lateness, and may respond with decreased service quality e.g., dropping of frames while displaying a video. The system to be developed is safety critical but at the same time a certain time span has to be provided to the algorithm to make sure whether the driver is really fatigued or the data obtained was just a one off glitch. The top– down mode of design is the most common approach to embedded system development. The major steps for this approach are:

i. Requirement
ii. Specification
iii. Architecture
iv. Components
v. System Integration

**The requirement of the system are given below**

a. It should be mountable in the car.

b. The system should be non intrusive, it should not cause any distraction to the driver

c. The false alarm rate should be a minimum

d. The system should operate in real-time

e. It should work for both day and nighttime.

**Requirement analysis**

Requirement analysis is done on the basis of the requirement of the final product. The set of specifications needed are decided from the requirement of the final product. The basic requirement for our system is given below.

*Table* 2.1 *System requirements*

| purpose | To monitor the alertness level of the driver using image based methods and to give an alarm if driver is drowsy |
|---|---|
| inputs(s) | Live Video from camera containing drivers face |
| output(s) | Alarm in case driver is drowsy |
| Functions | Detect the alertness level of driver from the camera video and to give an alarm in case driver is drowsy |
| Performance | The requirement for processing speed is a minimum of 4 frames per second. The accuracy of eye detection and classification should be higher than 90%. |
| Size | It should be compact enough to mount in the car. |
| Power | It should be able to draw power from car battery |
| Cost | Between 15000Rs-25000Rs |

**Specification**

The specification is a carefully documented technical statement which accurately reflects the customer's requirements in such a way, which can be clearly followed in a design. The Universal Modelling language (UML) is the most common approach to capture all these design tasks [15]. It is a standardized general purpose modelling language which uses graphical notation techniques to create models of specific systems. UML diagrams represent two different views of a system model.

Static (or structural) view: Emphasizes the static structure of the system using objects, attributes, operations and relationships. This approach consists of as many as 6 different types of diagrams.

Dynamic (or behavioural) view: Emphasizes the dynamic behaviour of the system by showing collaborations among objects and changes to the internal states of objects. This method consists of 7 different types of diagrams.

For our system, the best description would be behavioural state machine type diagram as we are more interested in the behavioural model of the system. Finally the algorithm is to be implemented on a development board. Therefore, structural diagrams are not considered though a rough idea of system resources required is necessary for board selection. Fig 2.1 shows the UML diagram of the system.

**Behavioural representation (state machine diagram in UML)**

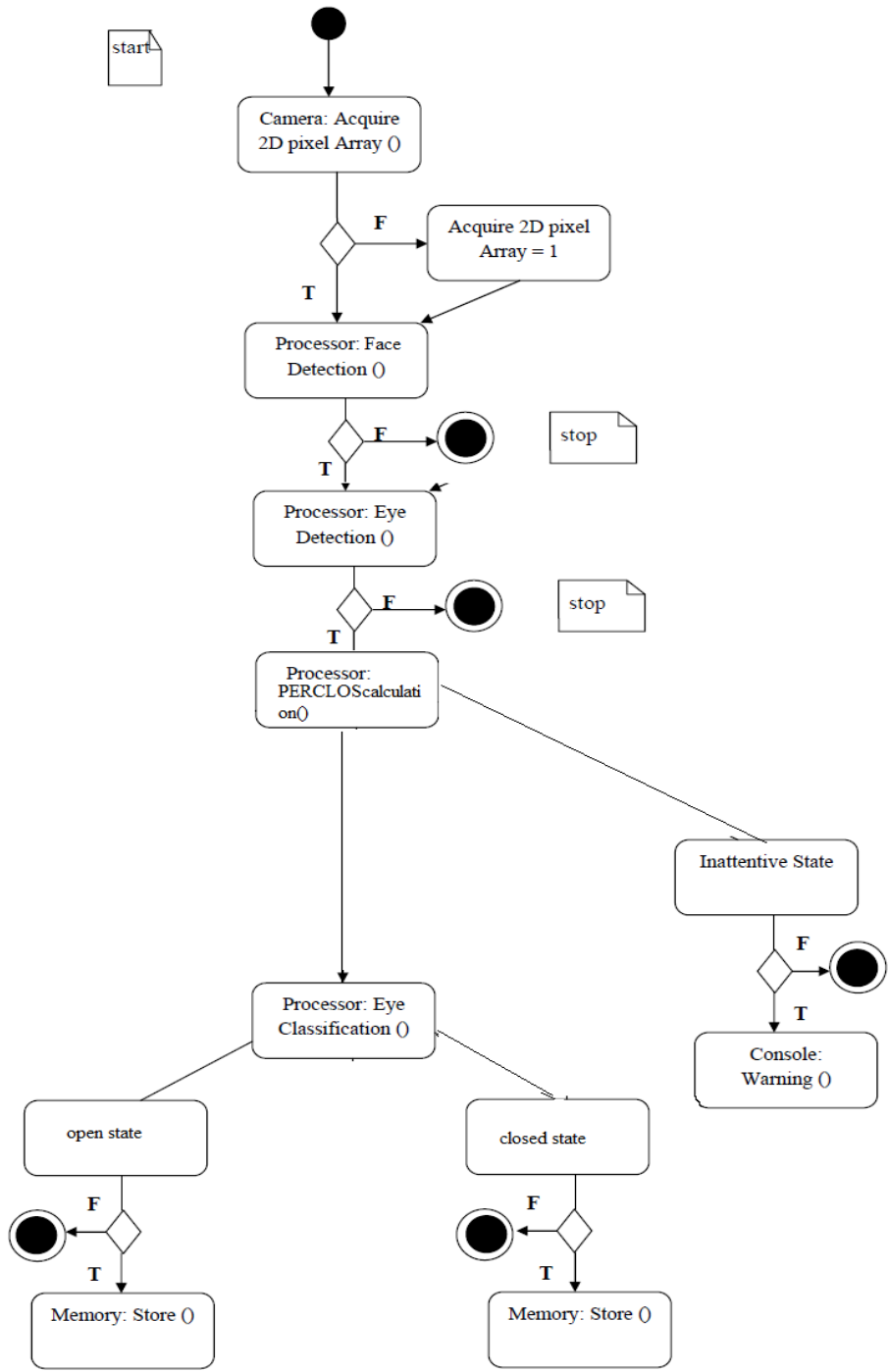

*Fig 2.1 State machine diagram in UML for the system*

**Block Diagram of the system**

**Hardware**

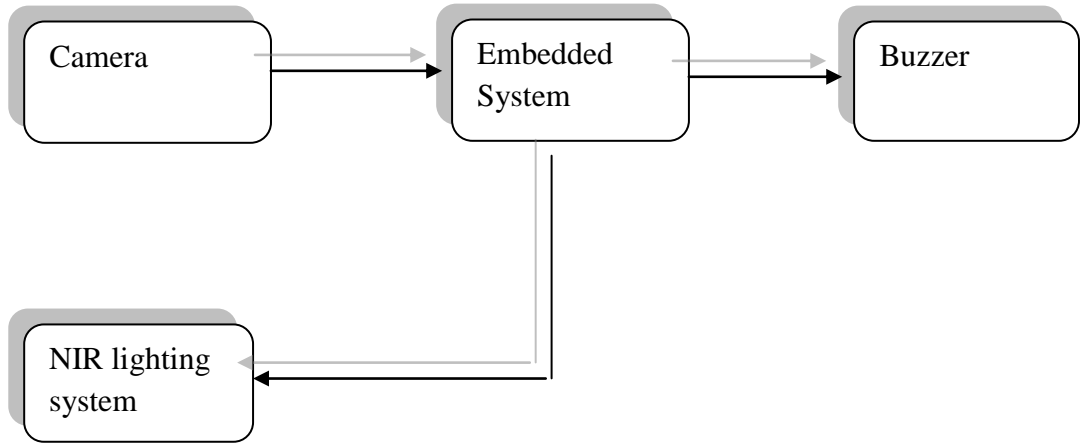

*Fig 2.2 Hardware block diagram of the system*

**Software**

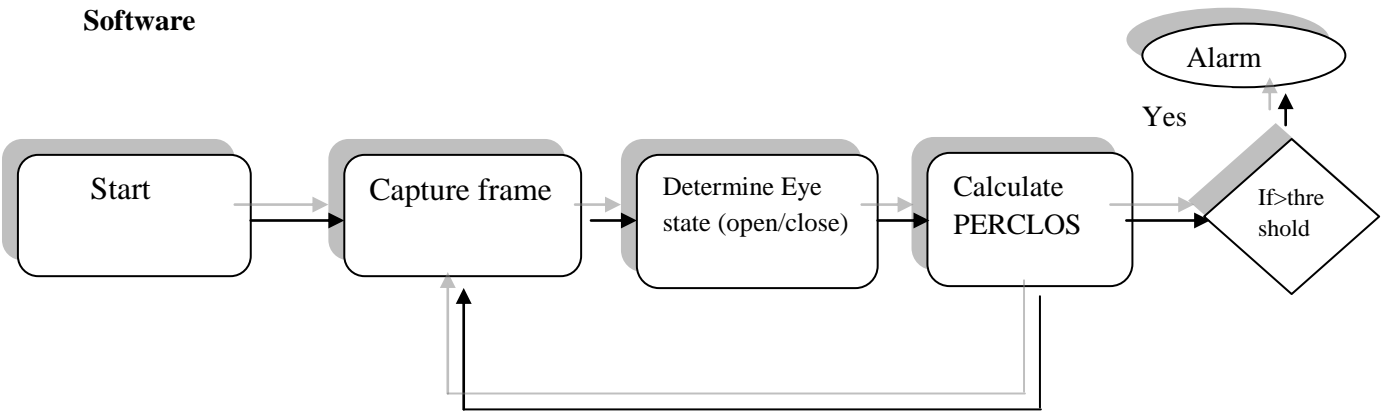

*Fig 2.3 Software block diagram of the system*

Fig 2.2 shows the overall hardware block diagram of the system. Camera and the embedded processing unit is the main part of the system. Fig 2.3 shows algorithm steps used for PERCLOS measurement. The loop runs continuously and the PERCLOS values are determined in overlapped time window of one minute.

**Selection of hardware platform**

We have considered two embedded development boards for the implementation , they are

a. Android 6410 development board

b. Intel atom N 450 board

a. **Android 6410 development board**

The Android6410 Single Board Computer is developed based on Samsung ARM11, which is designed to provide cost-effective, low-power capabilities, high performance Application Processor solution for consumer electronics, industrial control, GPS, industry pda etc.. Android6410 perfectly realized the performance of Samsung S3C6410 in video media decoding, 2d the 3d graphics, display and then put etc..It provides LCD interface, TV Out interface, camera input port, Serial ports, SD Card interface, SPI, 100m network Ethernet rj45 interface, USB2. 0 - OTG interface, USB host interface, audio input and output interfaces, i2c interface etc. it contains a higher CPU Frequency and richer peripheral, and can apply to the embedded situation which needs higher performance and processing capacity.

Supports four operating system WinCE, Ubuntu, Linux, Android
CPU frequency 667MHz
DDR 128M/256M Bytes DDR1
Nand Flash 256M Bytes

The main drawbacks of the system were

1. The lower CPU speed

2. Lack of proper documentation for cross compilation of libraries and drivers for camera.

b. **Intel atom N 450 board**

Intel Atom board is having a Intel Atom N450 processor, with a cpu speed of 1.66GHz, 1GB DDR2RAM, VGA port, 6USB ports, 2 Ethernet ports, Serial port and a DIO port. It can be powered from a DC source of 12V.

We have selected this platform for implementation owing to its higher processing speed and compatibility withX86 architecture. The board can work on 12 V DC supply so it could be directly connected to the car battery

**Intel Atom Specifications**

i. Processor- Intel Atom N450, 1.6GHz
ii. RAM-1 GB single channel DDR2 667 MHz
iii. Ethernet- 2 Ethernet ports
iv. VGA port-1
v. USB- 6 (USB 2.0 compatible)
vi. Audio-3( Mic -in, Line-in, Line-out)
vii. Serial port- 1 RS 232
viii. DIO- 8 bit GPIO
ix. Operating system support- Windows Embedded XP, Linux, CE6.0
x. Power requirements- DC 12V input (Tolerance ±10%)
xi. Operating temperature- 0-60° C
xii. Dimensions –255x152x50 mm
xiii. Weight- 2.5 Kg

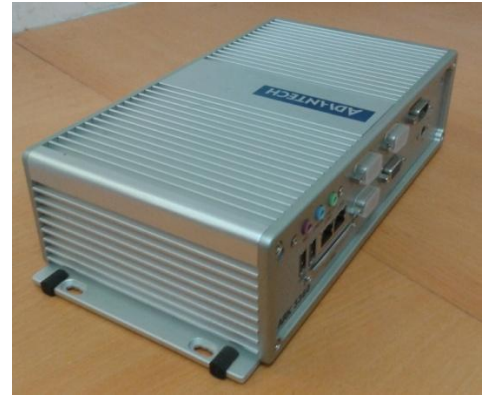

*Fig* 2.4. *Intel Atom Board*

The Intel atom board selected for our application is shown in Fig 2.4

**Camera Hardware**

The main aspect of the camera to be used include good resolution image. It should be able to transmit the videos at 30 fps through USB. During night time NIR illumination is used so the camera should be sensitive in NIR region (750-2000nm). For our application we have selected an I Ball Webcam with a resolution of 640x480 with maximum frame rate 60.

**Camera Positioning**

The placement of camera should be selected in accordance with four constraints

- ✓ It should not obstruct the view of the driver
- ✓ The image acquired should contain driver face at its centre
- ✓ Direct light from other vehicles or street lights should not fall on the camera
- ✓ Effect of vehicle vibration should be minimum

Based on the above constraints the camera placement is done on top of the dashboard. A hood is to be given to reduce the effect of light directly falling on the camera. On the final implementation the size of the camera will be very small so that it can be concealed in the dashboard.

**Lighting system**

During night time we use an active NIR illuminator for lighting up the face. We have used 24 NIR LED's for the NIR detector. The illuminator positioning has been done in such a way that it gives maximum illumination to the face. Fig 2.5 shows one part of the NIR lighting system.

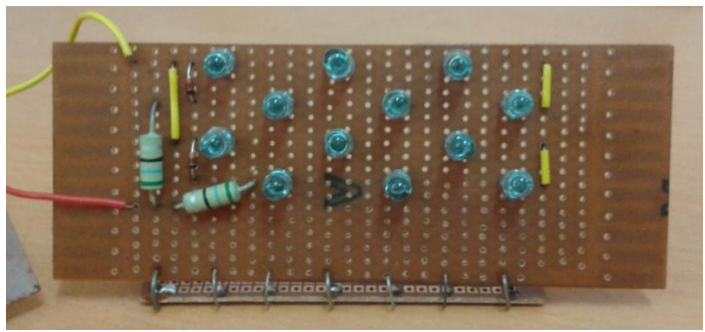

*Fig 2.5 NIR lighting system*

**Power**

The system is to be powered from car battery. Use of extra battery will need frequent maintenance. We have used the 12V direct outlet from the car for powering the system. The main issue with the usage of battery is the surges during starting. We have used voltage stabilization to prevent surges.

**Software Platform**

Real time performance is the key requirement for the selection of software platform. The software platforms we have tested are Windows XP Embedded, Windows CE, Ubuntu and Android. The software platform selected for our system is Windows Embedded XP. The IDE used for development is Visual Studio 2006.

*Pictorial view*

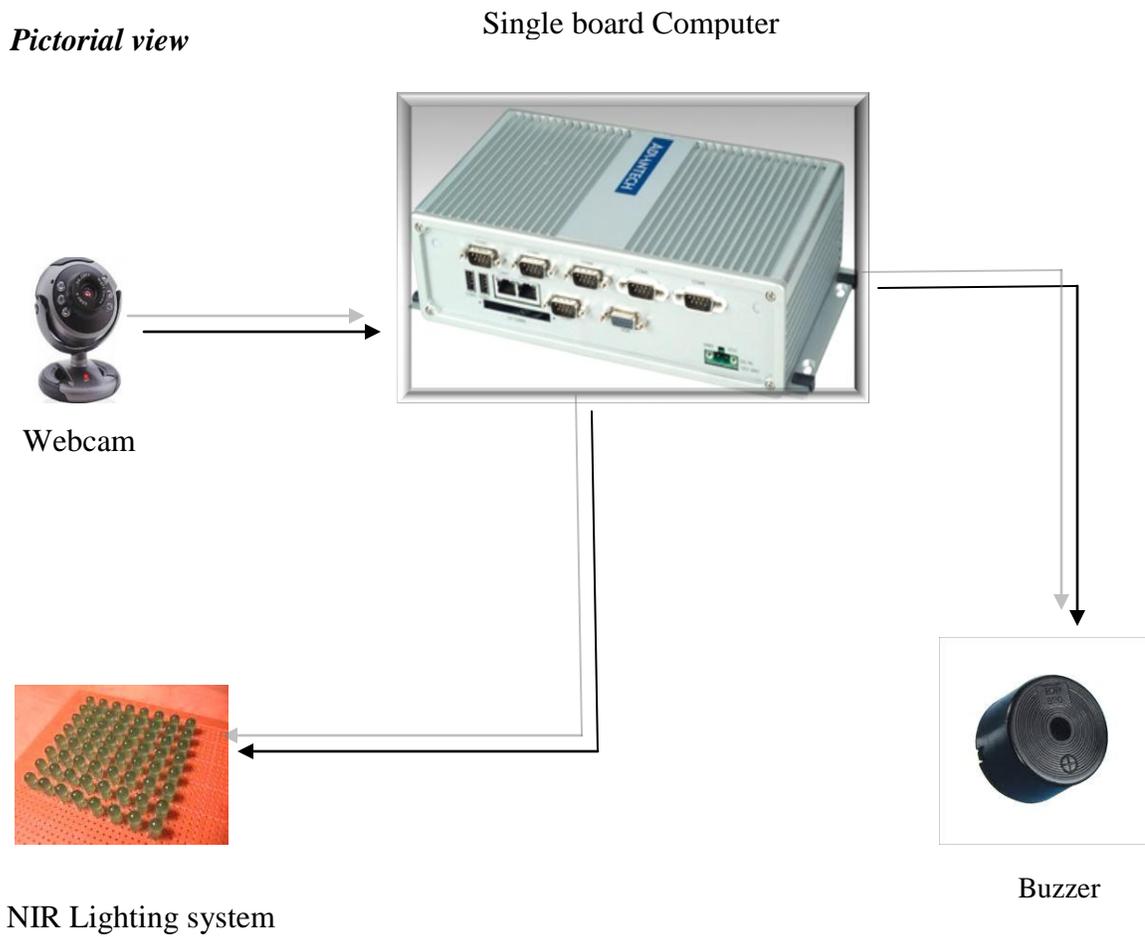

*Fig* 2.6 *Pictorial view of the system*

Fig 2.6 depicts the arrangement of parts in the final system. USB web camera is used as the input device. The NIR lighting system is another part. The buzzer is used to generate alarm when PERCLOS value is more than a threshold.

# Chapter 3

# FACE DETECTION AND EYE LOCALIZATION

## 3.1. Introduction to haar like features

Primarily Haar classifiers have been used for Face and Eye detection. A rectangular Haar-like feature can be defined as the difference of the sum of pixels of areas inside the rectangle, which can be at any position and scale within the original image. Each Haar like feature consists of two or three jointed black and white rectangles. The Haar wavelets are a natural set basis functions which encode differences in average intensities in different regions. The value of a Haar-like feature is the difference between the sums of the pixel gray level values within the black and white rectangular regions:

$f(x) = Sum_{black\ rectangle}\ (pixel\ gray\ level) - Sum_{white\ rectangle}\ (pixel\ gray\ level)$ (3.1)

The advantage of using Haar like features over raw pixel values is that it can reduce/increase the in-class/out-of-class variability, which makes the classification easy. Fig 3.1 shows the commonly used rectangular features.

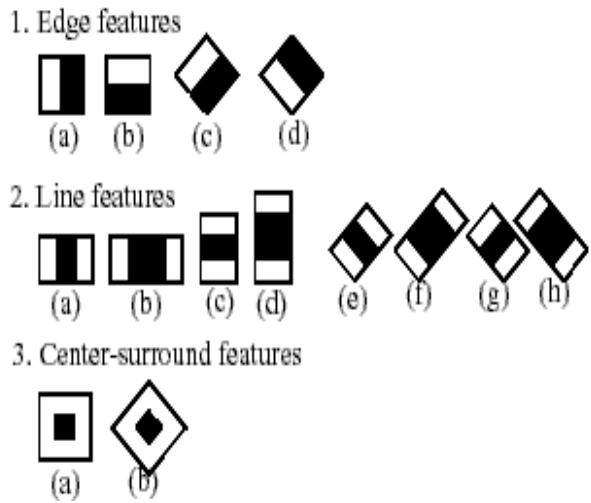

*Fig* 3.1 *Common rectangular features*

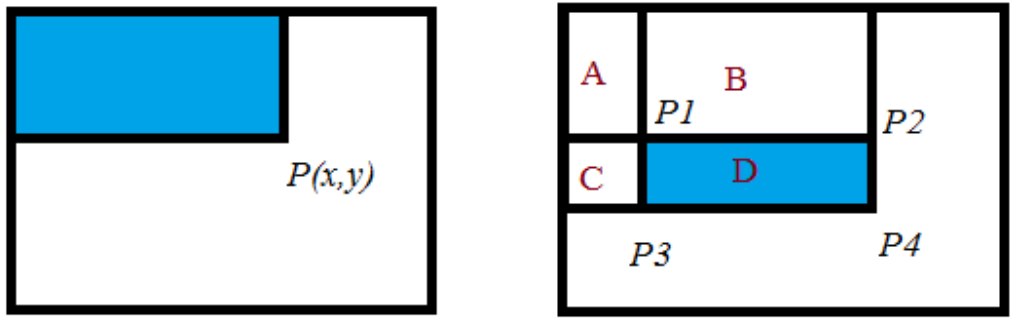

*Fig 3.2 Integral image representation*

From Fig 3.2, Integral image at location of x, y contains the sum of the pixel values above and left of x, y, inclusive:

$$P(x,y) = \sum_{x'<x, y'<y} i(x', y') \quad (3.2)$$

$$P_1 = A, P_2 = A + B, P_3 = A + C, P_4 = A + B + C + D \quad (3.3)$$

$$P_1 + P_4 - P_2 - P_3 = A + A + B + C + D - A - B - A - C = D \quad (3.4)$$

The computation of Haar like features is made fast by the use of integral image representation. The number of Haar like features in an image is too large. In a sub image of size 24x24 the number of features contained is 180,000.The Haar like features are evaluated in a cascaded manner .Weak classifiers reject the region where the probability of finding face is less. More time is spent on the promising regions in the image .The features used in each stage and their threshold is selected in training phase. Training is done with AdaBoost algorithm [9]. In training phase the classifier is trained with a number of positive and negative samples to get a cascaded classifier. It consists of cascaded stages of weak classifiers [9]. The optimal set of cascaded features and corresponding thresholds are obtained from the AdaBoost algorithm.

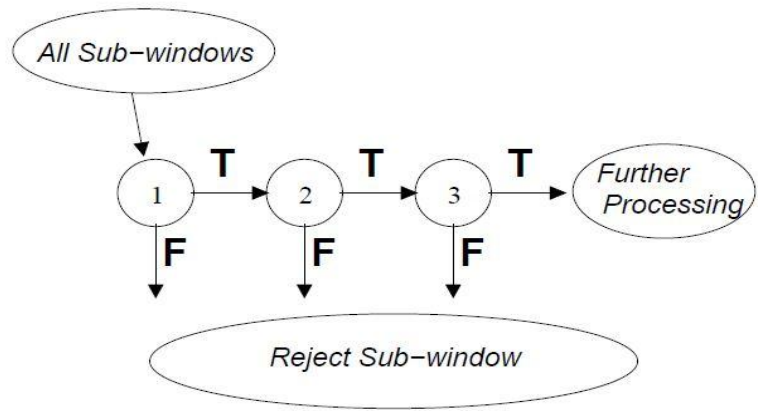

*Fig* 3.3 *Cascade structure*

A series of classifiers are applied to every sub-window. The cascaded structure is shown in Fig 3.3. The first classifier eliminates a large number of negative images and passes almost all the positive sub-windows (high false positive rate). The subsequent stages of the cascaded detectors reject the negative sub-windows passed by the first stage weak classifier. After several stages of processing the number of negative windows will be reduced greatly

**3.2 .Application Of Haar classifier in face detection**

The frames obtained from camera are of resolution 640x480. First Haar cascade for detection of face is used. The cascaded stages are applied on the image in every scale and at every location. Once the face is detected a region of interest (ROI) is selected from the face region, based on face geometry. Now in this ROI Haar cascade for closed eyes is applied if it detects a closed eye a counter increments and camera fetches next frame and it is processed. If closed eye is not detected Haar classifier for open eyes are checked. After it is complete it takes the next frame from camera and processes it. In each minute the PERCLOS value is calculated as the ratio between number of closed eyes detected and number of eyes found. From the threshold of PERCLOS, fatigue level decision is taken, and can be used to alarm the driver.

### 3.3. Modified algorithm for high frame rate

In the above explained algorithm the full resolution picture is examined to detect the face. Integral images are used to calculate features rapidly in many scales. Integral image can be calculated by a few operations per pixel. Once computed, any one of these Harr-like features can be computed at any scale or location in *constant* time [9], from this it is apparent that after the integral image has been calculated the time needed for Haar feature detection is the same. So in order to speed up the detection time the time taken for the calculation of integral images must be minimized.

The integral image at location x, y contains the sum of the pixels above and to the left of x, y inclusive:

$$ii(x, y) = \sum_{x' \leq x, y' \leq y} i(x', y') \tag{3.5}$$

The time needed to calculate the integral image increases with the size of the image. In order to decrease the time taken, the image is down sampled. The number of pixels in the new image will be reduced. This improves the speed of detection of face. Haar classifier used for face performed very well for scaling factor up to 7, with original image of resolution 640x480.

One disadvantage of this method is that, it reduces the detection rate when the scale factor is more than 6. So a trade off between speed and accuracy was made by setting the scale factor to 6. It gives almost the same detection rate as the original image.

From the face region detected from the down sampled image, a region of interest for the detection of eye is selected. To detect the eye state with maximum accuracy we need the region of interest (ROI) at the same resolution as the image captured from camera. This is done by remapping ROI co ordinates to original image.

**Remapping ROI co-ordinates to original image**

We have used a method where the coordinates of ROI obtained from scaled down image is used for obtaining the region to perform eye detection. A pictorial representation of the schemeis given in Fig 3.4.

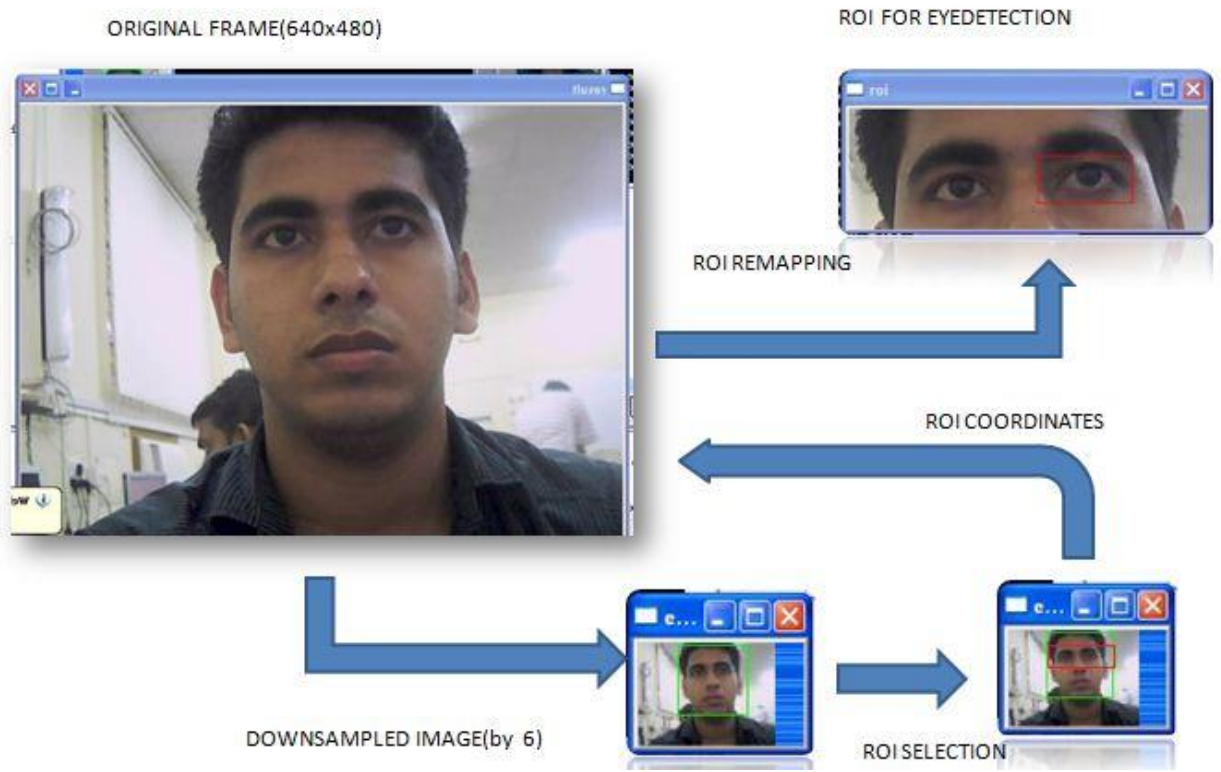

*Fig 3.4 Remapping ROI co ordinates to original image*

The resolution of the acquired image (N) is 640x480. It is down sampled by a scale factor '*x*'. The new image '*D*' obtained is having a resolution $((640/x) \times (480/x))$.
Face detection is carried out in the down sampled image and co-ordinates of bounding rectangle of face is obtained $(x_1, y_1)$, $(x_2, y_2)$, $(x_3, y_3)$ and $(x_4, y_4)$
All the co-ordinates are multiplied by scale factor '*x*' to get a new set of co ordinates

($x.x_1,x.y_1$), ($x.x_2,x.y_2$), ($x..x_3,x.y_3$) and ($x.x_4,x.y_4$) these co ordinates are mapped into the original image 'N'

From the mapped region in 'N', a rectangle region is selected which is having a distance ($h/5$) from top of face and ($h/3$) from bottom of face, where '$h$' is the height of bounding rectangle of face in 'N'

This ROI is resized to a resolution of 200x70. It is used for eye detection in subsequent stages. This method ensures the very same detection rate of eyes as in the original algorithm and at the same time, improving the execution speed considerably.

The results of frame rate with different scaling factors are shown in the table below

*Table 3.1 Scale factor Vs Speed*

| Scaling factor | Frames per second |
| --- | --- |
| 1 | 2.5 |
| 2 | 5 |
| 4 | 7.5 |
| 6 | 9 |
| 8 | 10 |

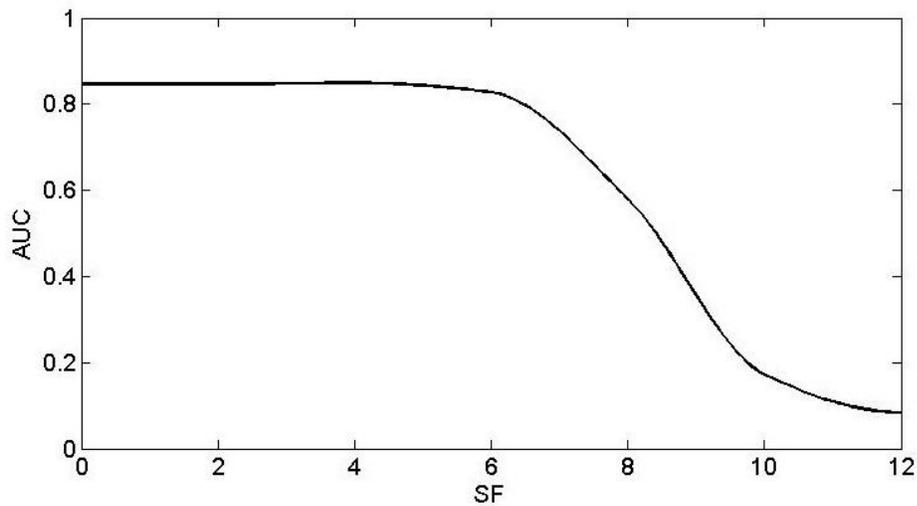

*Fig 3.5 Scale factor Vs Area Under Curve (AUC)*

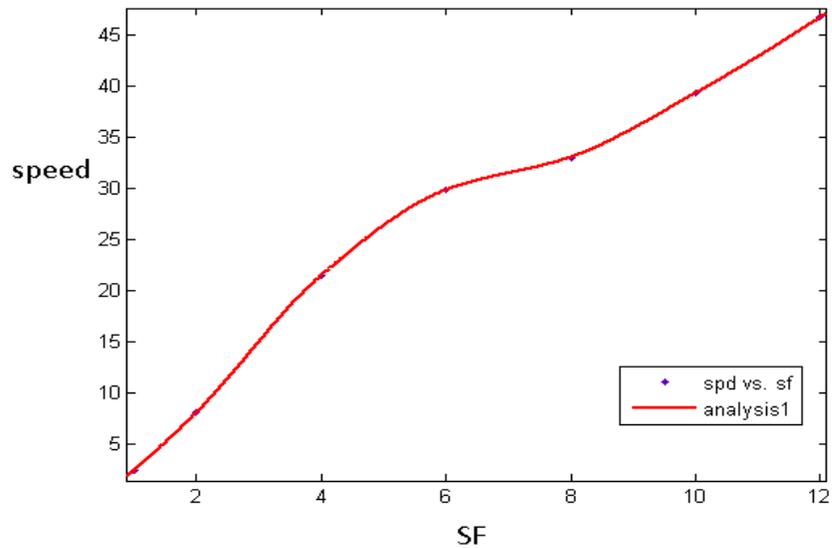

*Fig* 3.6 *Scale factor Vs frame rate*

The speed versus SF analysis (Fig 3.6) reveals the fact that the speed of operation increases non-linearly with SF. This observation can be explained as follows. With down sampling of images, the number of pixels to be operated on reduces by a factor which is equal to the square of SF. This reduces the time needed to calculate the integral images. Once the integral images are computed, the time needed for Haar-like feature based detection at any scale and location is constant. Further, the number of sub windows, and number of scales of search also reduced which in turn improves the speed even more. Hence, such an increase in speed is observed with increase in down sampling SF. The accuracy versus SF is shown in Fig 3.5. From this it is apparent that the accuracy and AUC remains almost constant up to SF of 6 and then drops down up to an SF of 10 and saturates henceforth. From these observations scale factor of 6 is selected as a trade off between speed and accuracy.

## 3.4. Tilted face detection with affine transform

The original Haar cascade technique applied for face detection detects upright face only. If there is a moderate amount of tilt of face it will not be detected, consequently eyes will also be not detected in such frames. Since the angle in which driver faces camera can vary with driving conditions, it is necessary to detect tilted face.

Some initial approach in tilted face detection used tilted face for training. Its accuracy was found to be very less. Defining tilted rectangular features is another method proposed by Lienhart and Maydt [16]. There are other methods that estimate the pose of face in a sub image and de rotates the window. If it is a non-face, conventional classifier will reject this window in the second stage. But here final detection rate is low (depends on accuracy of both classifiers).

We have implemented a method based on affine transforms.

**Affine Transform**

An affine transformation is an important class of linear 2-D geometric transformations which maps variables (*e.g.* pixel intensity values located at position (*x,y*) in an input image) into new variables (*e.g.* $x_2, y_2$ in an output image) by applying a linear combination of translation, rotation, scaling and/or shearing (*i.e.* non-uniform scaling in some directions) operations. The advantage of using affine transformation is that it preserves collinearity(ie the points lying on a line will lie on a line even after transformation) and ratio of distances(the mid point of line segment will remain the midpoint of the transformed image). These two properties assures that the affine transformed faces will be detected by the Haar classifier.

Vector algebra uses matrix multiplication to represent linear transformations, and vector addition to represent translations. Using an augmented_matrix and an augmented vector, it is possible to represent both using a single matrix multiplication. The technique requires that all vectors are augmented with a "1" at the end, and all matrices are augmented with an extra row of zeros at the bottom, an extra column—the translation vector—to the right, and a "1" in the lower right corner. If *A* is a matrix,

$$\begin{bmatrix} \vec{y} \\ 1 \end{bmatrix} = \begin{bmatrix} A & \vec{b} \\ 0,\dots,0 & 1 \end{bmatrix} \begin{bmatrix} \vec{x} \\ 1 \end{bmatrix} \quad (3.6)$$

is equivalent to the following

$$\vec{y} = A\vec{x} + b \quad (3.7)$$

In 2D graphics, for rotation by an angle θ **counter clockwise** about the origin,

Written in matrix form, as:

$$\begin{bmatrix} x' \\ y' \end{bmatrix} = \begin{bmatrix} cos\theta & -sin\theta \\ sin\theta & cos\theta \end{bmatrix} \begin{bmatrix} x \\ y \end{bmatrix} \quad (3.8)$$

The rotation matrix can be found for an "*n*" dimensional image once its size, center and angle of rotation needed are known. This is implemented in the algorithm.

**Steps Used to detect tilted face**

First Haar classifier for face is applied. If face is not detected then entire downscaled image is transformed clock wise and counter clockwise. After the detection the co ordinates of face are mapped to original image after de rotation. This eye localization method gives de rotated eye region. It makes the further processing more accurate. These steps improves the angle of tilt for face detection more than 70 degree, and It returns the ROI for eye detection in a de- rotated form , it further enhances the eye detection rate. Fig 3.7 shows the result of tilted face detection.

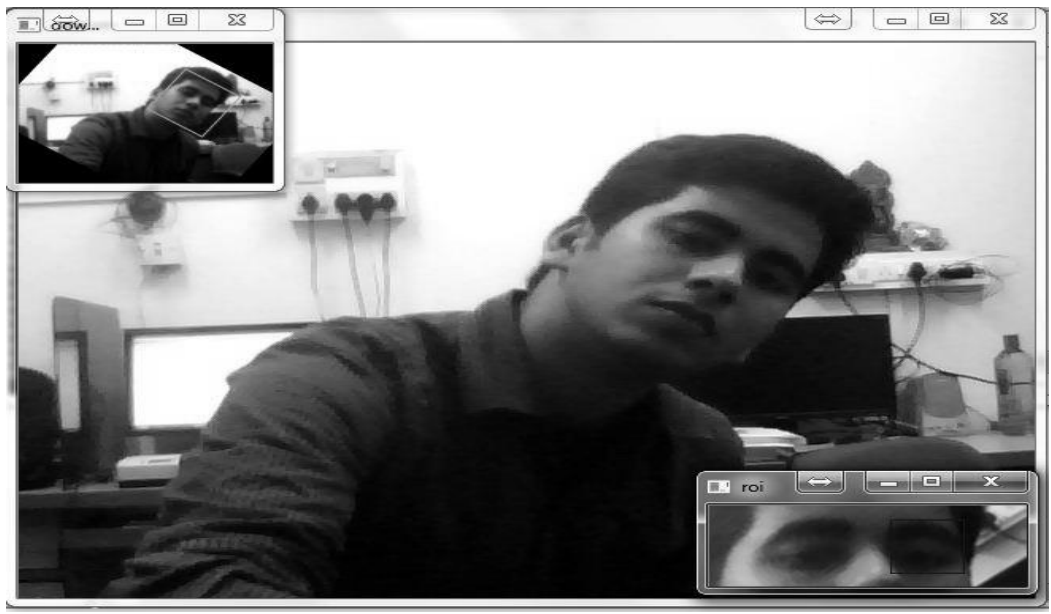

*Fig 3.7 Tilted face detection result*

## 3.5. Face tracking

Searching for face in every frame in every scale increases the computational complexity. The real-time performance of the algorithm can be improved if we use the temporal information. If the position and size of face is known accurately in a frame, then we can select a Region of Interest around that position where we can find the face in subsequent frames. The computational complexity is less since the search region is reduced.

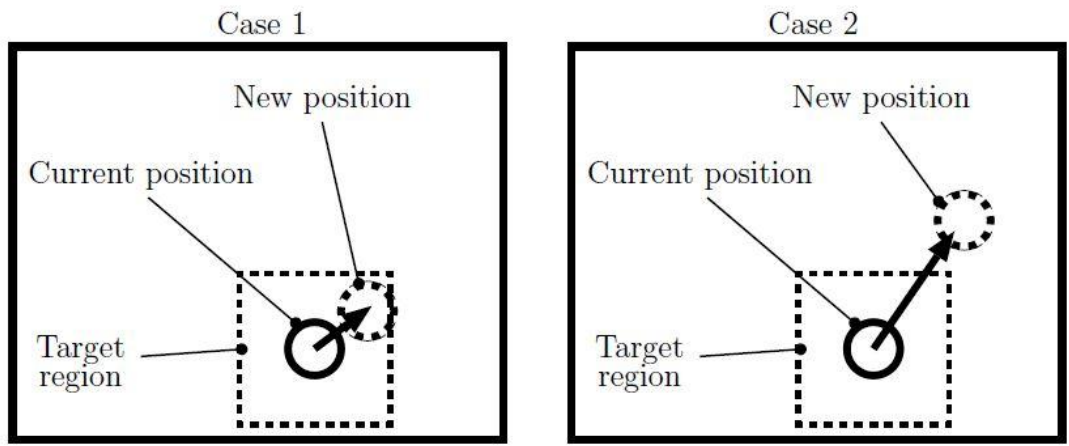

*Fig* 3.8 *Tracking in video*

Fig 3.8 shows two cases in tracking, if the position of face is accurately known in one frame, the position of the face can be predicted in subsequent fames. This reduces the computational complexity. But in the second case the motion of face is fast so the position of face will be outside the selected ROI. By tracking face with a prediction method avoid this problem. In tracking, the ROI is selected around the predicted position of face in the subsequent frame thereby reducing the tracking failure. There are several methods available for object tracking mean shift tracking, optical tracking and Kalman filter based tracking. The first two methods relies image intensity values in subsequent frames. The performance of mean shift tracking and optical tracking is poor because of the changing light conditions. We have selected Kalman filter based estimation since the prediction method depends on the measurement of face location from Haar based face detector only.

**Kalman Filter based tracker**

Kalman filter based tracker is used in the algorithm. The extreme ends of the bounding rectangle of face are taken as the measurement vectors. The extreme edge coordinates along with the velocity components in *x* and *y* direction is taken as the state vectors. Kalman filter predicts the position of bounding box of face. Now we select a region of interest around the predicted position to reduce the search space. When face is detected the co–ordinates of the bounding rectangle are used to update the Kalman filter [17]. When face is not detected the tracker reinitializes itself.

*Prediction of bounding rectangle:*

1. Predicted (a priori) state:
$$\hat{X}_{t|t-1} = F_t \hat{X}_{t-1|t-1} + B_t u_t \tag{3.9}$$

2. Predicted (a priori) estimate covariance:
$$P_{t|t-1} = F_t P_{t-1|t-1} F_t^T + Q_t \tag{3.10}$$

*Updating from observed measurements*

3. Innovation or measurement residual:
$$\widetilde{y_t} = z_t - H_t \tag{3.11}$$

4. Innovation (or residual) covariance:
$$S_t = H_t P_{t|t-1} H_t^T + R_t \tag{3.12}$$

5. Optimal Kalman gain
$$K_t = P_{t|t-1} H_t^T S_t^{-1} \tag{3.13}$$

6. Updated (a posteriori) state estimate:
$$\hat{X}_{t|t} = \hat{X}_{t|t-1} + K_t \tilde{y}_t \tag{3.14}$$

7. Updated (a posteriori) Estimate covariance:
$$P_{t|t} = (I - K_t H_t) P_{t|t-1} \tag{3.15}$$

Where,

$X_t$   is the current state vector as estimated by Kalman filter at time $t$

$Z_t$   is the measurement taken at time $t$.(left and right extreme co ordinates of the bounding rectangle)

$P_t$  measures the estimated accuracy of $X_t$ at time $t$

$F$  describes how system moves from one state to next (ideally), with no noise

$H$  describes the mapping from state vector $X_t$ to measurement vector $Z_t$

$Q$ and $R$   represent Gaussian process and measurement noise respectively and characterize     variance of the system

$B$ and $u$ are control input parameters are used in systems that have an input; these can be ignored in the case of face tracker.

**Results with Kalman filter**

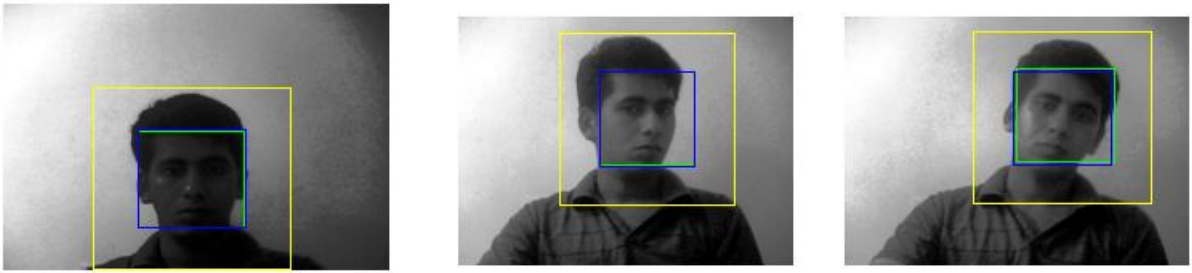

*Fig 3.9 Search window and estimated position when face is detected by Haar detector*

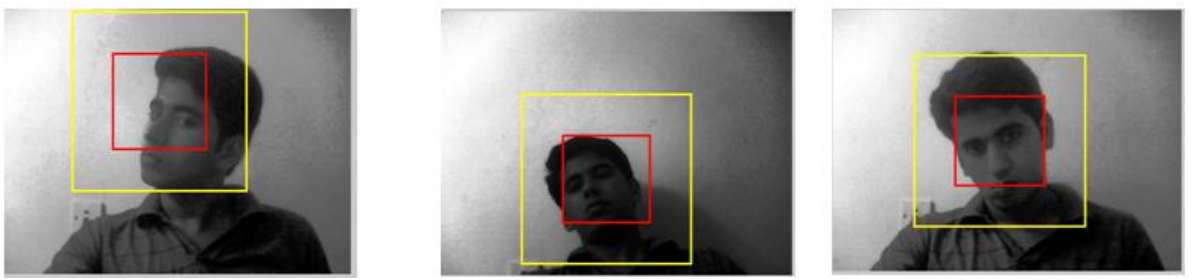

*Fig 3.10 Search window and estimated position when face is not detected*

In Fig 3.9 When face is detected in a frame, the search region for face in the next frame is reduced to a smaller region. In Fig 3.10 face is not found the prediction from Kalman filter can be used to reduce the search space for the subsequent stage of eye detection.

# Chapter 4

# EYE DETECTION

Once face is localized next step is to detect the position of eye. Subsequently the eye detected is classified to open or close. The detection of eye in the face region is modelled as an object detection problem. We have used two approaches for detection of eye in daytime. They are Haar classifier based eye detection and PCA based eye detection. Both of these methods act on grayscale images.

## 4.1. Haar classifier based eye detection

Two separate classifiers for open eye and closed eye are used in this method. The classifier for open and close are trained with a database and positive and negative images are given for training. The ROI selection is done and the detection of eye is performed in the localized region. The flow chart of the Haar based eye detection is shown here. The scheme used is depicted in Fig 4.1.

The number of open and closed eyes over one minute windows are calculated and PERCLOS values are found.

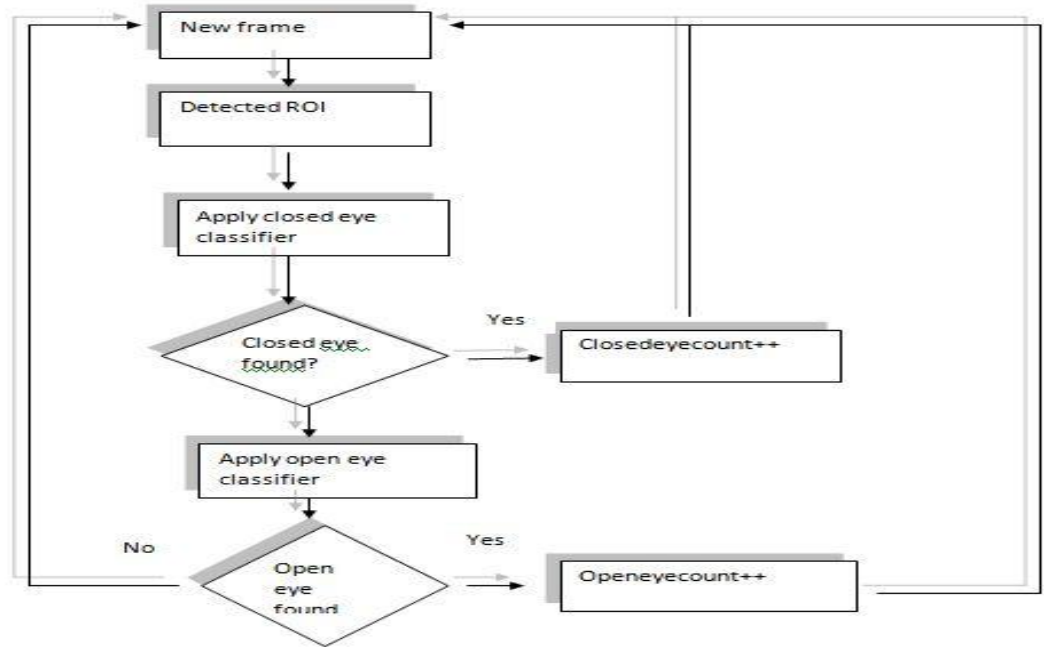

*Fig* 4.1 *Flowchart of eye detection with Haar classifier*

### 4.2. Principal component analysis

In Principal component analysis method each sub window is projected into PCA subspace and the sub window which is having minimum reconstruction error is selected as a positive detection. The size of eye window considered is 50x40, so the dimensionality of each vector is 2000. This increases the computational complexity. PCA subspace is used to reduce the computational complexity by projecting the data into a low dimensional subspace.

### 4.2.1 Introduction to Principal component analysis

PCA allows us to compute a linear_transformation that maps data from a high dimensional space to a lower dimensional sub-space. PCA uses an orthogonal transformation to convert a set of observations of possibly correlated variables into a set of values of uncorrelated variables called principal components [10]. Principal components are the directions in which the data varies the most. These directions are determined by the eigenvectors of the covariance matrix corresponding to the largest Eigen values. The magnitude of the Eigen values corresponds to the variance of the data along the eigenvector directions. The number of principal components would be less than that of original variables, there by reducing computational complexity. The sub images are projected to the principal components and energy components in each Eigen directions are found. The classification between open and closed eye is done based on the Euclidean distance of the sub images from the two classes. The sample is classified to a particular class if the Euclidean distance to that class is less than a threshold.

### 4.2.2 Algorithm

**PCA training algorithm**

1. For one class, obtain eye images $I_1, I_2, I_3 \ldots I_P$ (for training), each of dimension say $N \times M$

2. Represent every image $I_i$ of that class as a vector $\Gamma_i$ (of dimension N*M x1)

3. Compute the average eye vector

$$\psi = \frac{1}{P} \sum_{i=1}^{P} \Gamma_i \qquad (4.1)$$

4. Subtract the mean eye from each image vector $\Gamma_i$

$$\phi_i = \Gamma_i - \psi \tag{4.2}$$

5. since the covariance matrix $C$ given by:

$$C = \frac{1}{P}\sum \phi_n \phi_n^T = AA^T \tag{4.3}$$

Where $A = [\phi_1, \phi_2... \phi_P]$ ($N*M \times P$ matrix) is very large, Compute $A^TA$ ($P \times P$) instead as $P \ll N$

6. Compute the eigenvectors $v_i$ of $A^TA$. Using the equation

$$\sigma_i u_i = Av_i \tag{4.4}$$

Eigenvectors $u_i$ of $A A^T$ are obtained

7. Depending on computational capacity available, keep only $K$ eigenvectors corresponding to the $K$ largest Eigen values. These $K$ eigenvectors are the Eigen eyes corresponding to the set of $M$ eye images

8. Normalize these $K$ eigenvectors

9. Repeat the above procedure for the two classes of eye images

10. Given a test image $\Gamma$, subtract the mean image of each class:

$$\phi_j = \Gamma - \psi_j , (j = 1, 2) \tag{4.5}$$

11. Compute

$$\widehat{\phi_j} = \sum_1^k w_i\, u_i \text{ for each class } (j = 1, 2) \tag{4.6}$$

12. Compute

$$e_j = \|\phi - \widehat{\phi_j}\| \tag{4.7}$$

13. Classify test image as belonging to class j for which the norm $e_j$ is minimum

### 4.2.3. Application in image processing

In image processing the dimensionality of data is very large. The 2D images are first transformed to a vector by row ordering. The algorithm is trained with a number of training images and principal components are found. In the detection stage, each sub-window is projected to the principal components and the sub-windows for which the Euclidean distance is less than a threshold will be taken as a positive detection.

**The implementation of training phase PCA for eye detection includes the following steps**

1. Preparation of training set of eye region. It is taken from a video containing face .The pictures constituting the training set have been taken under the same lighting conditions,. It is histogram equalized to avoid light dependency .Each of the images is resized to same resolution to overcome scale dependency. Each training image is treated as one vector, by concatenating the rows of pixels in the original image, resulting in a single row with $m \times n$ elements ($m$- images, $n$-total pixels in a training image). For this implementation, it is assumed that all images of the training set are stored in a single matrix **T**, where each row of the matrix is an image.
2. Subtract the mean. The average image *a* has to be calculated and then subtracted from each original image in **T**.
3. Calculate the eigenvectors and Eigen values of the covariance matrix **S**. Each eigenvector has the same dimensionality (number of components) as the original training images, and thus can itself be seen as an image. The eigenvectors of this covariance matrix are therefore called Eigen eyes. They are the directions in which the images differ from the mean image.
4. Principal components are selected from the covariance matrix. The $D \times D$ covariance matrix will result in $D$ eigenvectors, each representing a direction in the $m \times n$-dimensional image space. The eigenvectors (Eigen eyes) with largest associated Eigen value are kept.

**Algorithm used to implement PCA eye detection**

1. Detect face with modified haar classifier algorithm with affine transforms. This gives a de rotated ROI as the output .This step reduces the false positives (searching along all sub windows is time consuming and gives rise to false positives)

2. Training data is loaded

3. The region of interest (ROI) is resized to match with the resolution of training images

4. Resized image is projected into the PCA subspace

5. The training set which is nearer to the projected image is found by using Euclidean distance.

6. The image is classified to the training set, to which it is closest.

7. PERCLOS values are calculated per minute and fatigue levels are determined

### 4.2.4 *Results with PCA*

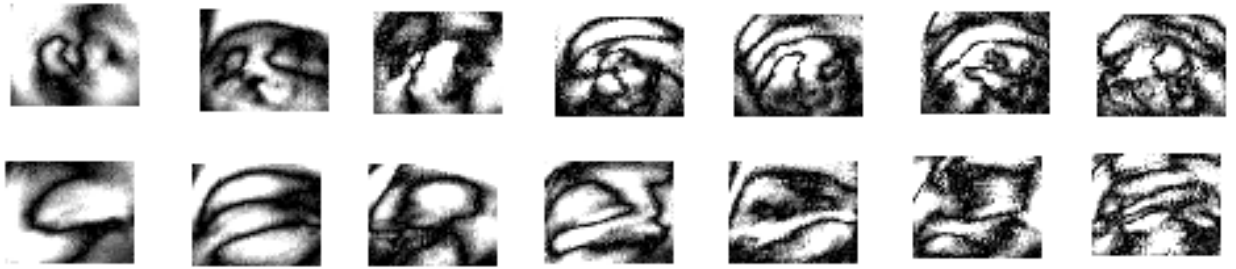

*Fig 4.2 Open and closed Eigen eyes*

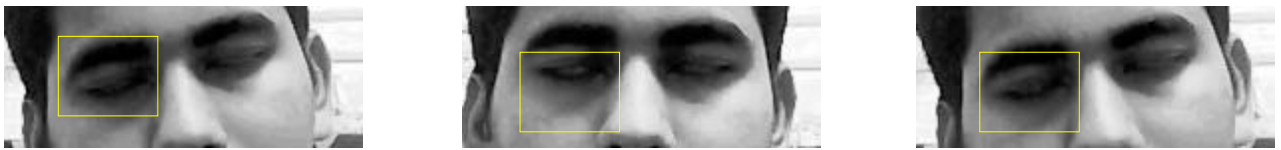

Closed eye detection

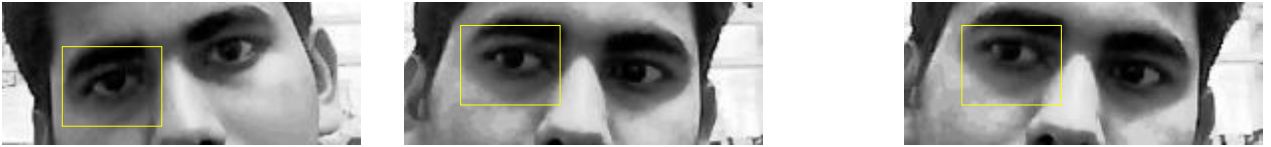

Open eye detection

*Fig 4.3 Eye detection results*

A test was conducted with 300 images 200 open and 100 closed eye frames. The face detection and ROI selection is done by Haar classifier based face detector. The detection results with PCA are shown in the table below. The accuracy rate was 98.5%. Fig 4.2 shows the open and closed Eigen eyes obtained from PCA algorithm. Fig 4.3 shows some of the detection results. Fig 4.4 shows the classification result from PCA.

*Table 4.1 Results with PCA*

| True Positive | True Negative | False Positive | False negative | True positive rate | False positive rate |
|---|---|---|---|---|---|
| 198 | 97 | 3 | 2 | 98.5% | 2.02% |

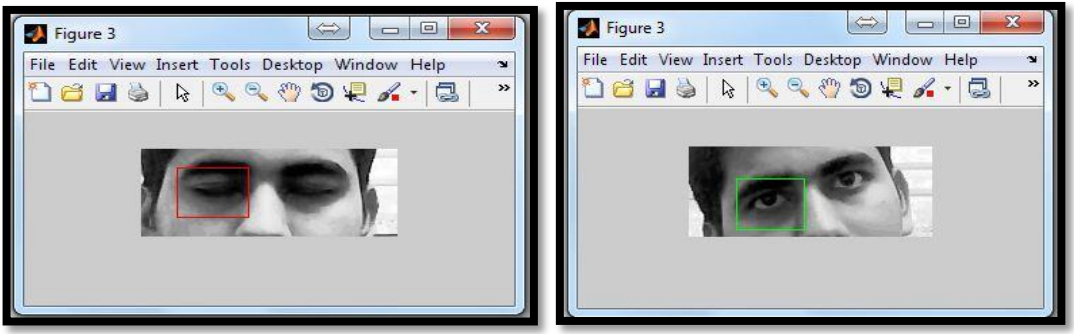

*Fig 4.4 Closed and open eye detection results*

# Chapter 5

## DETECTION OF EYE IN NIR IMAGES

Active NIR lighting is used to illuminate the face during night time. Haar classifier and PCA algorithms found to be giving poor results in detecting eyes with NIR images. The features are not prominent in NIR eye images. Block-LBP histogram features are used in detection of eyes in NIR images. Illumination invariance property of LBP feature is another advantage.

### 5.1. Local Binary Pattern (LBP) features

Local Binary pattern (LBP) features encodes the local features very efficiently. The local binary pattern (LBP) operator is defined as a gray-scale invariant texture measure, derived from a general definition of texture in a local neighbourhood. It labels the pixels of an image by thresholding the neighbourhood of each pixel and considers the result as a binary number. Due to its discriminative power and computational simplicity, LBP texture operator has become a popular approach in various applications. One of the main advantages of LBP is its illumination invariance [18].

If there is a texture T in a local neighbourhood of a grayscale image as the joint distribution of gray levels of P(P>1) image pixels.

$$T = t(g_c, g_0, g_1, \ldots\ldots, g_{p-1}) \tag{5.1}$$

Where a gray value $g_c$ corresponds to the gray value of the centre pixel of the local neighbourhood and $g_p$ $(P = 1,2,\ldots p-1)$ correspond to the gray value of P equally spaced pixels of a circle of radius

$$R(R > 0)$$

That form a circularly symmetric neighbour set.

If the coordinates of $g_c$ are (0,0) then the co ordinates of $g_p$ are given by

$$\left(-R\sin\left(\frac{2\pi P}{P}\right), R\cos\left(\frac{2\pi P}{P}\right)\right) \tag{5.2}$$

The Fig 5.1 shows circularly symmetric neighbour sets for various (P,R). The gray values of neighbours which do not fall exactly in the centre of pixels are estimated by interpolation.

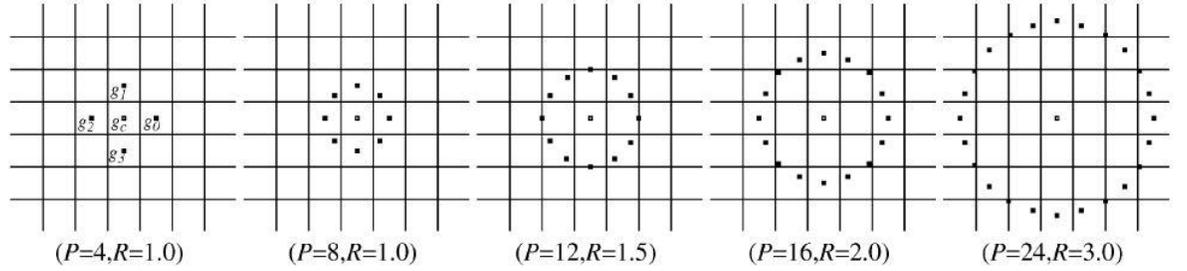

Fig 5.1 *Circularly symmetric neighbour sets for different (P, R)*

For grayscale invariance the gray value of the centre pixel ($g_c$) is subtracted from the circularly symmetric neighbourhood $g_p$ ($P = 1, 2, \ldots p-1$). Giving

$$T = t(g_c, g_0 - g_c, g_1 - g_c, \ldots, g_{p-1} - g_c) \tag{5.3}$$

Now we assume differences $g_p - g_c$ are independent of $g_c$, so

$$T = t(g_c) t(g_0 - g_c, g_1 - g_c, \ldots, g_{p-1} - g_c) \tag{5.4}$$

This is a highly discriminative texture operator. The occurrences of patterns in the neighbourhood of each pixel will be encoded in a P-dimensional histogram. In uniform regions the differences are zero in all directions. The Operator records highest difference in a gradient direction on a slowly sloped edge, Zero values along the edges and for a spot differences are high in all directions. The signed differences ($g_p - g_c$) are not affected by changes in mean illuminance: hence the joint difference distribution is invariant against grayscale shifts. The scale Invariance is obtained by considering the just the signs of the difference instead of their exact values.

$$LBP = \sum_{n=0}^{7} s(i_n - i_c) 2^n \tag{5.5}$$

Where,

$$s(x) = \begin{cases} 1, x \geq 0 \\ 0, x < 0. \end{cases}$$

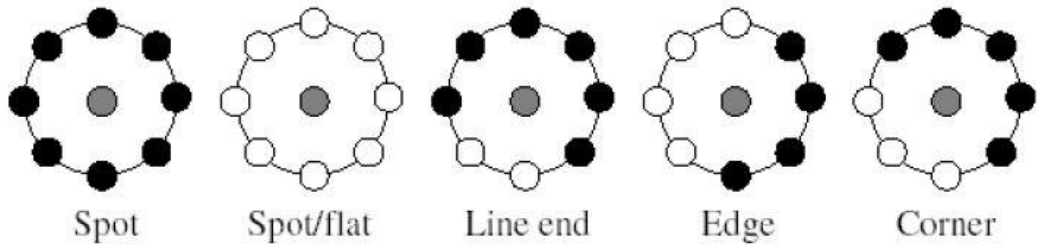

Spot    Spot/flat    Line end    Edge    Corner

*Fig 5.2 Example of texture primitives*

*Calculation of LBP feature*

| example | | | thresholded | | | weights | | |
|---|---|---|---|---|---|---|---|---|
| 6 | 5 | 2 | 1 | 0 | 0 | 1 | 2 | 4 |
| 7 | 6 | 1 | 1 |   | 0 | 128 |   | 8 |
| 9 | 8 | 7 | 1 | 1 | 1 | 64 | 32 | 16 |

Pattern = 11110001
LBP = 1 + 16 + 32 + 64 + 128 = 241

*Fig 5.3 Calculation of LBP*

Fig 5.3 shows the calculation of LBP feature, the steps for calculation of LBP feature is explained below.

- Divide each window to cells (e.g. 16x16 pixels for each cell).
- For each pixel in a cell, compare the pixel to each of its 8 neighbours (on its left-top, left-middle, left-bottom, right-top, etc.). Follow the pixels along a circle, i.e. clockwise or counter-clockwise.
- Where the centre pixel's value is greater than the neighbour, write "1". Otherwise, write "0". This gives an 8-digit binary number (which is usually converted to decimal for convenience).

- Compute the histogram, over the cell, of the frequency of each "number" occurring (i.e., each combination of which pixels are smaller and which are greater than the centre).
- Optionally normalize the histogram.

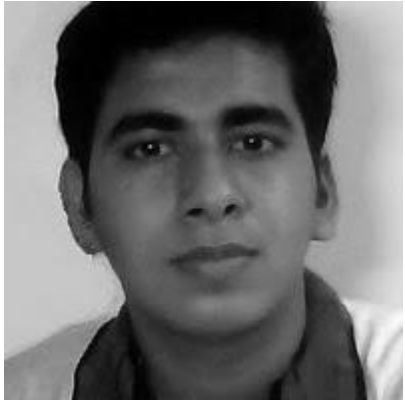      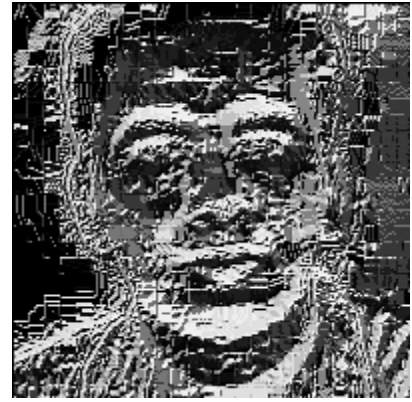

                Grayscale image　　　　　　　　　　　　LBP image

*Fig* 5.4 *Gray scale and corresponding LBP image*

Fig 5.4 shows a greyscale image and its corresponding LBP image.

### 5.2. Block-LBP histogram

The LBP algorithm encodes the local features efficiently. Block LBP histogram feature is having information at three levels global features, region features and the pixel level features.

In the LBP approach for texture classification, the occurrences of the LBP codes in an image are collected into a histogram. The classification is then performed by computing simple histogram similarities. But this results in loss of spatial information. One way to overcome this is the use of LBP texture descriptors to build several local descriptions of the eye and combine them into a global description. These local feature based methods are more robust against variations in pose or illumination than holistic methods.

In the LBP approach for texture classification, the occurrences of the LBP codes in an image are collected into a histogram. The classification is then performed by computing simple histogram similarities. However, considering a similar approach for facial image representation results in a loss of spatial information and therefore one should codify the texture information while retaining also their locations. One way to achieve this goal is to use the LBP texture descriptors to build several local descriptions of the face and combine them into a global description. Such local descriptions have been gaining interest lately which is understandable given the limitations of the holistic representations. These local feature based methods are more robust against variations in pose or illumination than holistic methods.

The basic methodology for LBP based face description proposed by Ahonen et al. (2006) [19] is as follows: The facial image is divided into local regions and LBP texture descriptors are extracted from each region independently. The descriptors are then concatenated to form a global description of the face, as shown in Fig 5.5.

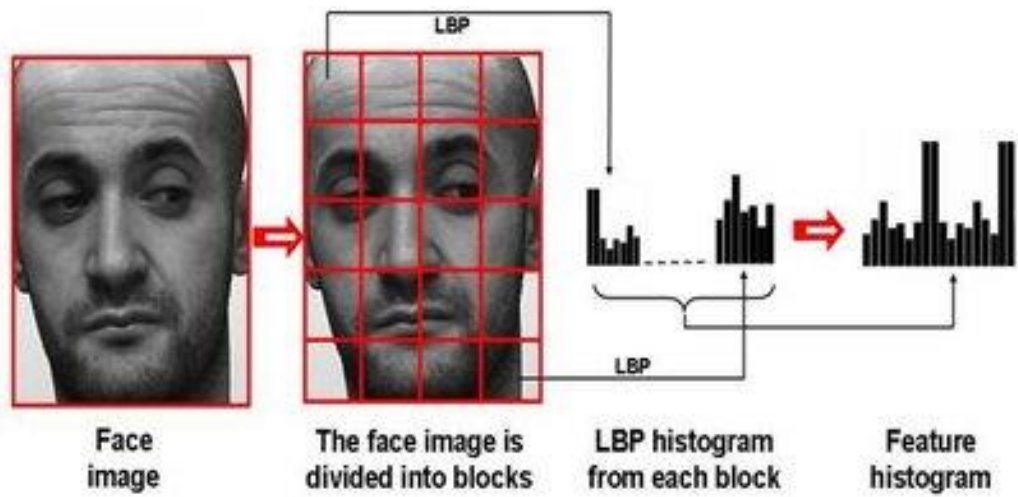

*Fig 5.5 Calculation of block LBP histogram*

### 5.3 *Algorithm*

Here we have used the local histograms and pixel level values to get the feature vector.

*Training Algorithm*

1. Open eyes of size 50x40 are selected and each image are divided in to sub-blocks of size 5x4
2. The LBP feature values of the sub-blocks are found
3. 16 bin histogram of each block is calculated
4. The histogram of each block is arranged to form a global feature descriptor
5. PCA is done on the feature vectors and 40 Eigen vectors having highest Eigen values were selected.

*Detection Phase*

1. ROI from face detection stage is obtained
2. For each sub window the Block- LBP histogram is found and is projected to Eigen space
3. The sub window with minimum reconstruction error is found
4. If the reconstruction error is less than a threshold it is considered as a positive detection

**Improvement of speed for calculation**

The speed of calculation for a single frame was prohibitive for a real-time application. The size of ROI from face detection stage is 200x70 and the size of eye for detection is 50x40. This means the number of sub windows to be searched is 4500. For each sub image the LBP feature is to be found and then the 16 bin multi block histogram. Once the feature vector is calculated it is projected to PCA subspace. In MATLAB the frame rate obtained was only 0.33. The number of search windows was limited by decreasing the overlap of pixels from 1 to 5 in horizontal and 4 in vertical direction. The calculation of LBP feature is done only once. From the Global LBP descriptor the 16 bin histograms are found at the first stage itself. Now in detection phase, the histograms are concatenated to give the description of the region. This approach reduced the computational to a great extent. In C coding the speed of the algorithm improved to a maximum of 6 fps.

**5.4 Results**

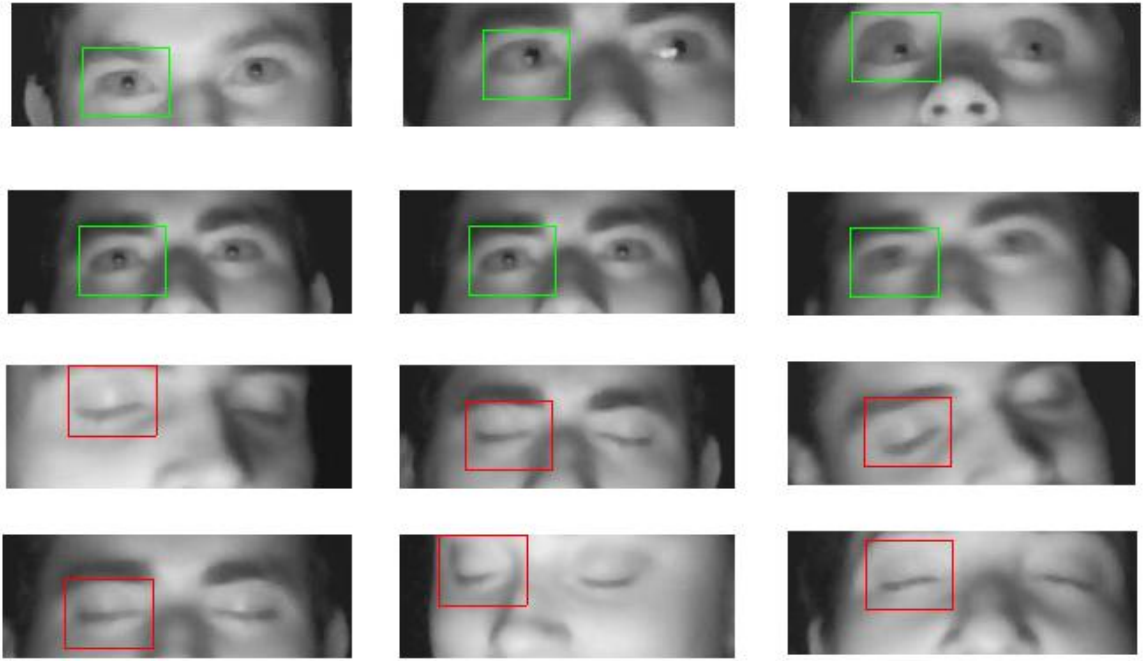

*Fig* 5.6 *Localization results with block LBP*

The Fig 5.6 shows the detection of eye in NIR image. The frame rate obtained was 6 fps. The algorithm was giving better results even with illumination variations.

# Chapter 6

## EYE STATE CLASSIFICATION

The eyes are detected by the block LBP histogram based approach. The final stage is the classification of state of the detected eyes. Support Vector Machine is reported as robust for the classification of open and closed eye [20]. The weight components from the previous stage are used for the classification of eyes.

**6.1. Introduction to Support Vector Machines (SVM)**

SVM is a supervised learning method applied for data classification. The standard SVM is a binary classifier. A support vector machine constructs a hyper plane or set of hyper planes in a high or infinite-dimensional space, which can be used for classification or regression. Good classification accuracy can be obtained if the hyper plane is maximally distant from the nearest training data from both the classes. When data cannot be classified by a linear classifier the original data can be transformed into a higher dimension where the classes can be separated by a hyper plane. The essence of SVM is to map the training data from the input space to a higher dimension feature space, where an optimal hyper plane can be found which can separate the data. In the SVM training process the dot product of input vectors have to be computed in the feature space [21]. The mapping of the input vectors to the feature space is achieved by using kernels, which actually directly computes the dot product of the input vectors in the feature space, instead of first transforming each input vector to the feature space and only then computing the dot product. There are various types of kernels available quadratic, polynomial, radial basis function (RBF) etc.

**6.2. Theory of SVM**

For a linearly separable data the principle of SVM is explained below. Fig 6.1 shows two linearly separable classes.
Given a set of linear separable training samples

$(\boldsymbol{x_i}, y_i)_{1 \leq i \leq N}, x_i \in R^d, y_i \in \{-1, 1\}$ (6.1)

$y_i$ is the class label where $x_i$ belongs to. The general form of linear classification function is[22]

$$g(x) = w.x + b \qquad (6.2)$$

This corresponds to a separating hyper plane $w.x + b = 0$

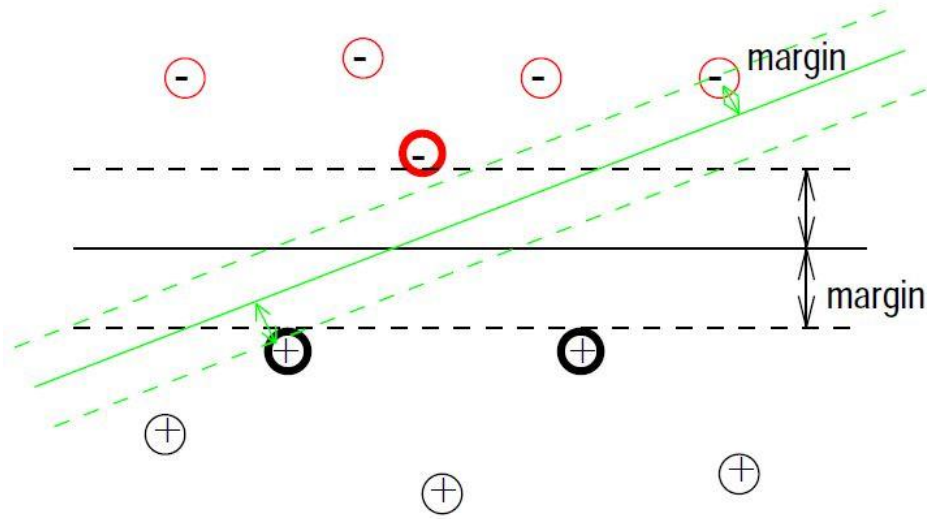

*Fig* 6.1 *Linearly separable classes*

We can normalize $g(x)$ to satisfy $|g(x)| \geq 1$ for all $x_i$, so that the distance from the closest point to the Hyper plane is $1/\|w\|$

Among the separating hyper planes, the one for which the distance to the closest point is maximal is called optimal separating hyper plane (OSH). Since the distance to the closest point is $1/\|w\|$ finding the OSH amounts to minimizing $\|w\|$ and the objective function is:

$$\min \emptyset(w) = \frac{1}{2}\|w\|^2 = \frac{1}{2}(w.w) \qquad (6.3)$$

Subject to

$$y_i(w.x_i + b) \geq 1, i = 1,2,..N \qquad (6.4)$$

If we denote by $(\alpha_1, \alpha_2, \ldots \alpha_N)$ the $N$ non-negative Lagrange multipliers associated with constraints in (6.3, 6.4), we can uniquely construct the OSH by solving a constrained quadratic programming problem. The solution $w$ has an expansion

$$w = \sum_i \alpha_i y_i x_i \tag{6.5}$$

In terms of a subset of training patterns, called support vectors, which lie on the margin. The classification function can thus be written as

$$f(x) = sign(\sum_i \alpha_i y_i x_i . x + b) \tag{6.6}$$

When the data is not linearly separable, on the one hand, SVM introduces slack variables and a penalty factor such that the objective function can be modified as

$$\phi(w,) = \frac{1}{2}(w.w) + C(\sum_1^N \xi_i) \tag{6.7}$$

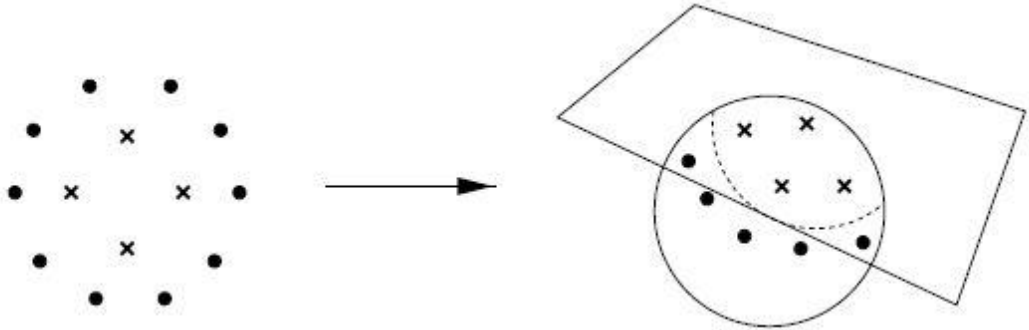

*Fig 6.2 Mapping to a higher dimension*

On the other hand, the input data can be mapped through some nonlinear mapping into a high-dimensional feature space in which the optimal separating hyper plane is constructed. Thus the dot product can be represented by

$$k(x, y) = (\phi(x).\phi(y)) \tag{6.8}$$

When the kernel $k$ satisfy Mercer's condition [23]. Fig 6.2 shows a mapping from 2 dimensions to 3 dimensions. Finally, we obtain the classification function

$$f(x) = sign(\sum_i \alpha_i y_i . k(x, x_i) + b) \tag{6.9}$$

## 6.3. Application

Here the eye state classification is modelled as a binary class problem. The class labels include open class and closed class. In the training phase from the open and closed eyes are feature transformed with Block-LBP and then projected to PCA subspace. The weights obtained are used for the training input of the SVM. The weight data along with the ground truth is used for the training. In the detection phase, the localized eye region is feature transformed first and then projected into the PCA feature space to get the weight components. The weights are fed into the SVM and SVM returns the classification result. From this eye state is obtained. The value of PERCLOS is found by calculating the ratio in overlapped time windows of one minutes and alarm is given when it is more than a threshold.

## 6.4. Results

The training of SVM has been carried out with 460 images and it's tested with another 1700 samples of data. The training images are loaded to MATLAB Then block LBP histogram feature transform is done. Then PCA is carried out on the feature vectors. Now the weight vectors corresponding to each sample is found by projecting the samples to PCA subspace. These weights along with the ground truth are used for SVM training. The training is done with different kernels and different accuracy levels were obtained. The results are shown below

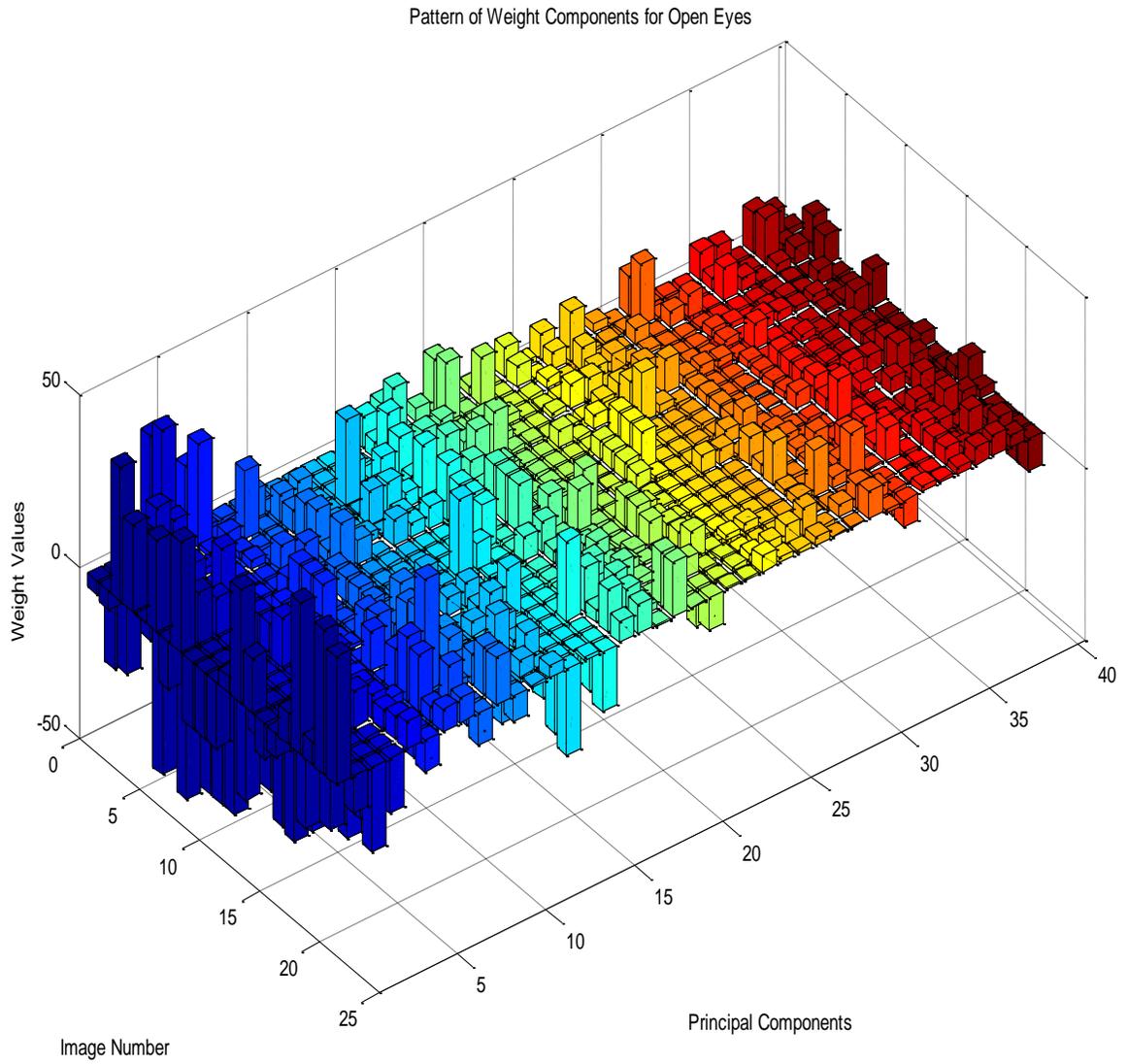

*Fig 6.3 Weight pattern for Open eyes*

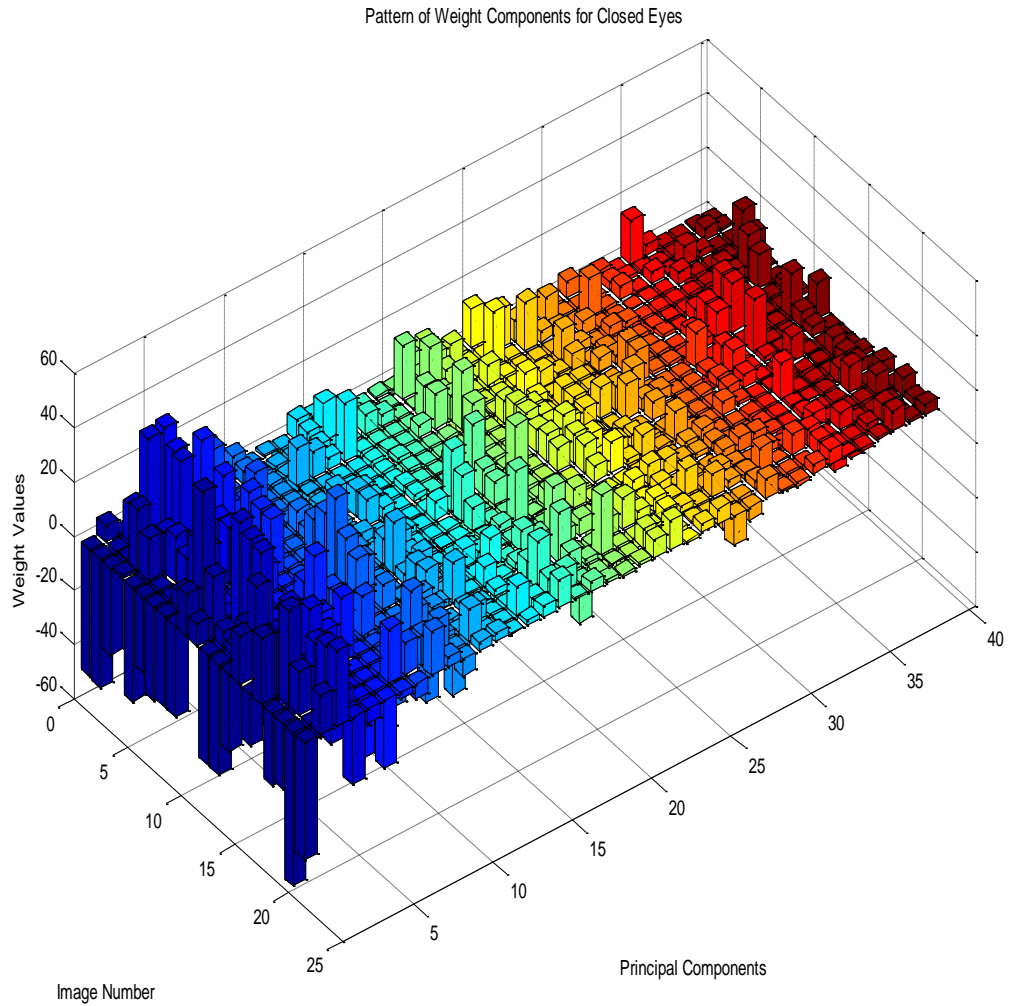

*Fig* 6.4 *Weight pattern for closed eyes*

Fig 6.3 and Fig 6.4 shows the weight patterns for open and closed eye images.20 images of each class are projected after the LBP feature transform. From the weight pattern of first 10 weight components the variation of weight components for open and closed classes is apparent. The table shows the detection results. SVM with polynomial kernel was found to give good accuracy. Fig 6.5 shows some samples of eyes used for the testing.

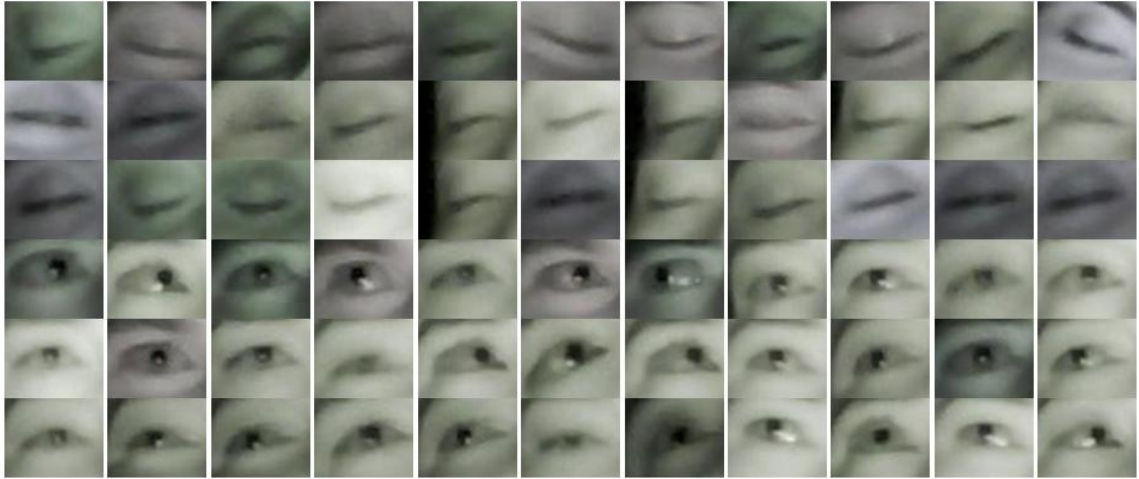

*Fig* 6.5 *Some samples of eyes used in the testing*

*Table* 6.1 *Detection results with SVM*

| Kernel function | True positive(tp) | True negative(tn) | False positive(fp) | False negative(fn) | True positive rate(tpr) | False positive rate(fpr) |
|---|---|---|---|---|---|---|
| Linear SVM | 838 | 808 | 42 | 12 | 98.58% | 4.94% |
| Quadratic | 827 | 765 | 85 | 23 | 97.29% | 10% |
| Polynomial | 848 | 807 | 43 | 2 | 99.76% | 5.32% |

# Chapter 7

## ON BOARD TESTING

An on board experiment was carried out to check the operation of the designed platform. One camera was placed directly on steering column just behind the steering wheel, and the other was towards the left of the steering wheel on the dashboard (Fig7.2 Fig 7.3) . The SBC was powered from the car battery. The pictures of the on board test are given below

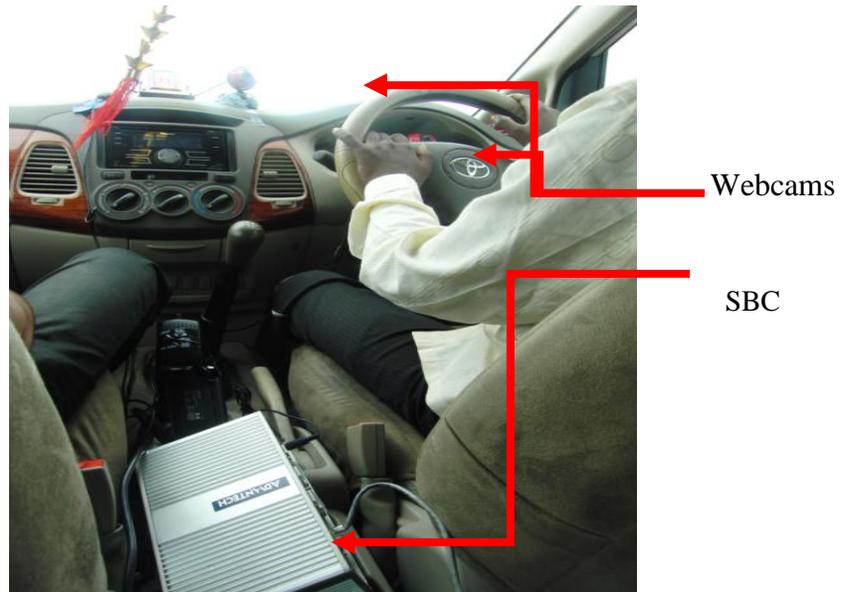

*Fig.* 7.1 *On board testing of the system*

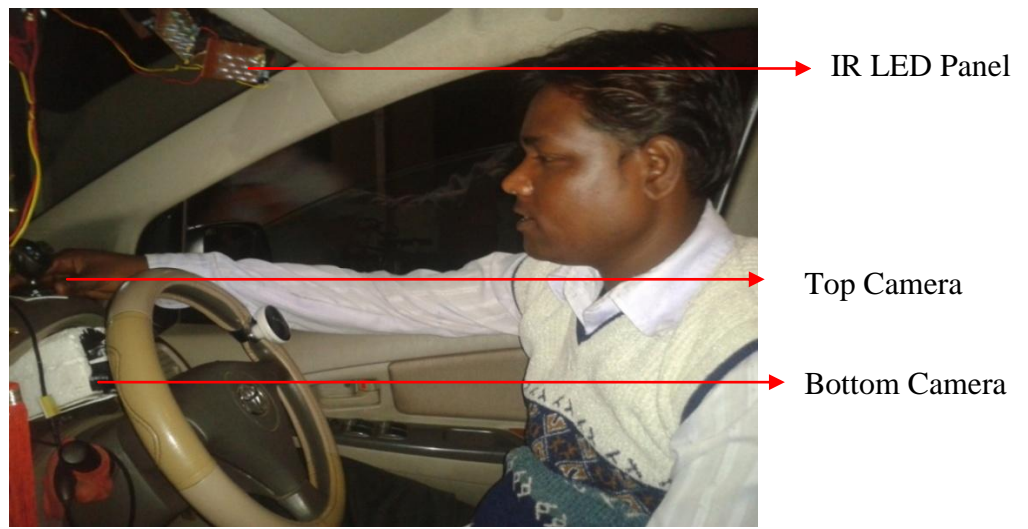

*Fig.* 7.2 *Arrangement of camera and lighting system*

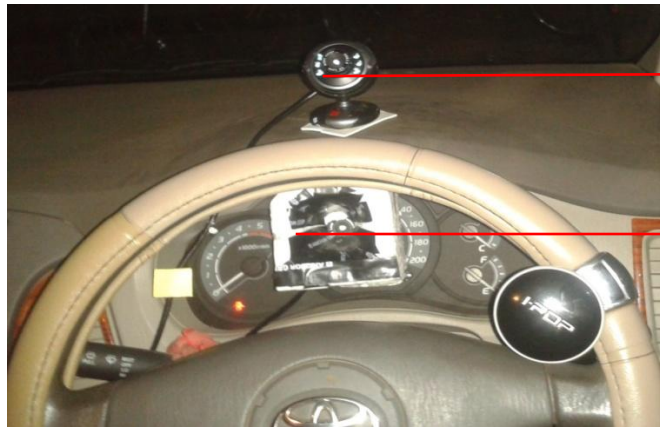

Camera on Top

Camera on Steering Wheel (mounted inside a thermocol hood to protect the camera lens from direct falling of light)

*Fig* 7.3 Camera placement

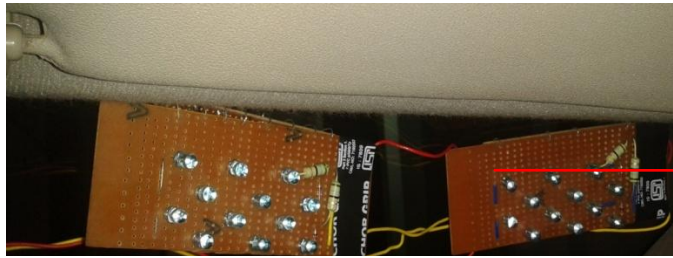

IR LED Lighting glued just beneath the ceil

*Fig* 7.4 NIR lighting system

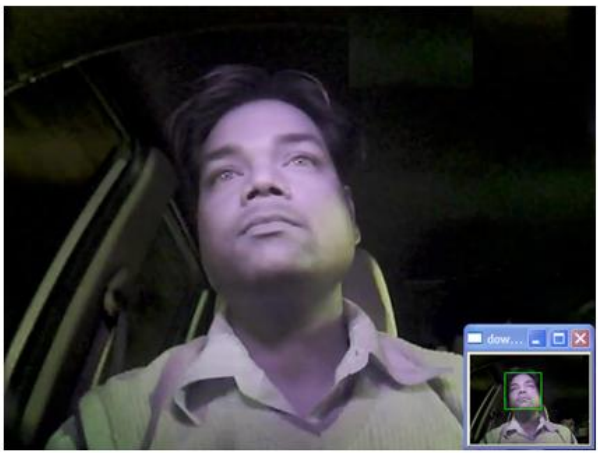

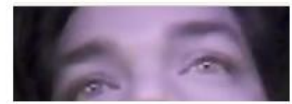

Detected ROI

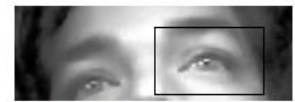

Detected Open Eye

*Fig. 7.5 Detection of eyes in NIR lighting*

NIR lighting system works well for the test duration. Lighting was enough to detect faces properly providing a clear NIR image as input to the SBC. Fig 7.5 shows the detection from the system. Detection of face and eye in NIR images was quite successful with satisfactory results.

# Chapter 8
# DISCUSSIONS AND CONCLUSIONS

Real time implementation of the algorithm has been done. This method has proven to be useful under laboratory as well as on board testing conditions with dark and lighted illumination conditions .The PERCLOS has already established to be an effective method by my co researchers at IIT Kharagpur [24]. The detection rate of the final algorithm is well above the required accuracy rate. The speed of computation is more than the requirements needed.

- ➢ The Haar classifier based algorithm for face detection was found to be working with accuracy more than 90% in daylight conditions as well as NIR lighting. The face detection method was working up to a frame rate of 9, and detects tilted faces with an in plane rotation of 70 degree.
- ➢ Performance of Haar based eye detection poor in NIR lighting, since the features available in NIR lighting is low
- ➢ PCA based algorithm performed with an accuracy of 98% in detection eyes in daytime.
- ➢ PCA can tolerate moderate amounts of tilt and can be modified to detect eyes with spectacles
- ➢ The performance of PCA depends on the training set, so a representative training set has to be selected for the training.
- ➢ LBP feature based methods gave best results in NIR eye localization
- ➢ The eye state classification accuracy from SVM was more than 98%, which was suitable enough for the application.

# Chapter 9

# FUTURE SCOPES

- The speed of the algorithm implemented is more than the required frame rate, so more computationally intensive post processing can be used to reduce the false positive rates.
- Tracking of face with Condensation algorithm can be used to detect the eyes even if the face detection stage fails.
- The detection of face in off plane rotation can be solved with the use of suitable tracker algorithms.
- A specially designed camera can be used which can be concealed in the dashboard, it will reduce the problem of light directly falling on to the camera
- The NIR lighting is to be designed for final implementation
- The packaging of the system is to be done for automotive conditions.
- The certification by ARAI (Automotive Research Association of India) has to be obtained.

# Chapter 10